\documentclass[3p,11pt]{elsarticle}

\usepackage{titlesec}

\usepackage{geometry}
\usepackage{graphicx}
\usepackage{setspace}

\newcommand{\w}{\omega}

\newcommand*{\red}{\textcolor{black}}
\newcommand*{\blue}{\textcolor{black}}

\newtheorem{observation}{Observation}

\newtheorem{proposition}{Proposition}
\newtheorem{lemma}{Lemma}
\usepackage{bm}
\usepackage{dsfont}
\usepackage{amssymb}
\usepackage{amsmath}
\usepackage{url}

\usepackage{algorithm}
\usepackage{algpseudocode}
\usepackage{setspace}

\usepackage{enumitem}
\usepackage{xcolor}
\definecolor{mygreen}{RGB}{0,0,0}

\usepackage{hyperref}
\usepackage{verbatim}
\usepackage{amsfonts}

\usepackage{bbm}
\usepackage{dsfont}
\usepackage{bm}
\usepackage{xcolor}
\usepackage{physics}

\usepackage{subcaption}
\usepackage{adjustbox}
\graphicspath{{images/}{../images/}}
\usepackage{rotating}
\usepackage{lscape}
\usepackage{blindtext}
\usepackage{enumitem}
\usepackage{subcaption}
\usepackage{graphicx}
\usepackage[graphicx]{realboxes}
\usepackage{hyperref}

\floatname{algorithm}{SMO Algorithm}

\usepackage{adjustbox}
\usepackage{tabularx}
\newtheorem{corollary}{Corollary}

\newcommand{\1}{\mathbf{1}}

\journal{Computers \& Operations Research}

\begin{document}

\begin{frontmatter}



\title{Binary Kernel Logistic Regression: a sparsity-inducing formulation and a convergent decomposition training algorithm} 

\author[mymainaddress]{Antonio Consolo\corref{mycorrespondingauthor}}

\author[mysecondaryaddress]{Andrea Manno}
\author[mymainaddress]{Edoardo Amaldi}
\cortext[mycorrespondingauthor]{Corresponding author}

\address[mymainaddress]{DEIB, Politecnico di Milano, Milano, Italy}
\address[mysecondaryaddress]{Centro di Eccellenza DEWS, DISIM, Università degli Studi dell’Aquila, L'Aquila, Italy}


\begin{abstract} Kernel logistic regression (KLR) is a widely used supervised learning method  
for binary and multi-class classification, which provides estimates of the conditional probabilities of class membership for the data points. Unlike other kernel methods such as Support Vector Machines (SVMs), KLRs are generally not sparse. \blue{Previous attempts to deal with sparsity in KLR include a heuristic method referred to as the Import Vector Machine (IVM) and ad hoc regularizations such as the $\ell_{1/2}$-based one. Achieving a good trade-off between prediction accuracy and sparsity is still a challenging issue with a potential significant impact from the application point of view. In this work, we revisit binary KLR and propose an extension of the training formulation proposed by Keerthi et al., which is able to induce sparsity in the trained model, while maintaining good} \textcolor{mygreen}{testing accuracy}.
To efficiently solve the dual of \textcolor{mygreen}{this} formulation, we devise a decomposition algorithm of Sequential Minimal Optimization type which exploits \textcolor{mygreen}{second-order} information, and
for which we establish global convergence.
Numerical experiments conducted on 12 datasets from the literature 
show that the proposed binary KLR approach achieves a competitive trade-off between accuracy and sparsity
with respect \textcolor{mygreen}{to IVM}, $\ell_{1/2}$-based regularization for KLR, and \textcolor{mygreen}{SVM} 
while retaining the advantages of providing informative estimates of the class \textcolor{black}{membership} probabilities.
\end{abstract}




\end{frontmatter}



\section{Introduction}

Logistic Regression (LR) is a popular supervised learning \textcolor{black}{method} in Machine Learning (ML) and Statistics, \textcolor{black}{which} holds significant relevance for classification and predictive analytics. 
It belongs to the class of generalized linear models, where the log-odds of an event are modeled as a linear combination of features. 
\textcolor{black}{Since LR models directly provide estimates of the 
conditional	probabilities that the data points belong to the classes, they}
are widely used in applications in which  
\textcolor{black}{this type of information is of interest. Indeed, in a variety of settings (e.g., credit risk scoring) the class \textcolor{mygreen}{membership probabilities turn} out to be even more useful than the class membership predictions.}
Examples of \textcolor{black}{LR applications include}: anomaly detection in data for fraud identification (e.g., \cite{mok2010random,EDGAR201795}), disease prediction in medical contexts (e.g., \cite{BAGLEY2001979,schober2021logistic}), and forecasting in business analytics (e.g., \cite{nikolic2013application,de2018new}). 

Kernel Logistic Regression (KLR) extends standard LR by using kernel functions to capture complex nonlinear relationships within \textcolor{black}{the} data (see e.g. \cite{jaakkola1999probabilistic}). In particular, in KLR the original input data is mapped into a higher-dimensional \textcolor{black}{(implicit)} feature space where metrics are computed using 
kernel functions. \textcolor{black}{The expressive power of the LR model is substantially increased by generating highly nonlinear decision 
boundaries with respect to the original input space, while preserving the} 
\textcolor{mygreen}{probabilistic nature of LR}. 

\textcolor{black}{During the last two decades, KLR has been applied 
\textcolor{mygreen}{in a} variety of fields. Early applications are related, for instance, to speaker recognition and rare event prediction. In \cite{katz2006sparse} the authors show that KLR and especially sparse KLR perform better in speaker identification experiments than standard Gaussian mixture models and Support Vector Machines (SVMs) on a benchmark dataset. In \cite{maalouf2011robust} the authors extend KLR by combining rare events corrections and truncated Newton methods, and demonstrate that, based on the statistical significance test, KLR is more accurate than SVM. Later, in \cite{hong2015spatial,chen2017gis} KLR \textcolor{mygreen}{is} 
used in geological hazard risk assessment to identify areas susceptible to landslides, showing a better overall performance compared to SVMs and decision trees. Recently, KLR \textcolor{mygreen}{has also been applied in discrete choice modeling}, which 
\textcolor{mygreen}{is} central to analysing individual decision making, with wide-ranging applications in economics, marketing and transportation. In \cite{martin2020discrete}, KLR is used to specify non-parametric utilities in Random Utility Models and is compared \textcolor{mygreen}{with} 
Multinomial Logit models in terms of goodness of fit and the capability of obtaining the specified utilities. In \cite{martin2021revisiting} a similar approach is adopted to model random utilities and empirical results show the advantages of KLR with respect to SVMs and Random Forests in terms of accuracy and of the estimation of important indicators in the field of transportation. Finally, in \cite{martin2025scalable} the authors introduce a Nyström approximation for KLR on large datasets in order to improve scalability and evaluate performance using large-scale transport mode choice datasets.}


\textcolor{mygreen}{As previously mentioned, the distinctive feature of binary KLR is to naturally provide probabilistic estimates of the class membership that are interpretable and are useful to estimate expected costs of classification decisions and confidence levels. It is worth pointing out that this is in contrast with other popular kernel methods such as binary SVM, which frequently achieve similar classification performance (see e.g. \cite{keerthi2005fast}). Since SVM predictions are deterministic, several approaches have been proposed to try to derive such valuable probabilistic class membership estimates from the SVM outputs. These include adopting a specific parametric family for the posterior probability \cite{wahba1999support}, using a series of trigonometric functions \cite{vapnik1998statistical}, applying logistic calibration \cite{platt1999probabilistic}, and employing sequential training of weighted classifiers to obtain an interval-based estimation \cite{wang2008probability}. More recently, several attempts have been made to either propose new alternatives (e.g., \cite{franc2011support,benitez2024cost}) or address the limitations of previous approaches (e.g., \cite{boken2021appropriateness,zeng2023sparse}). Despite the growing attention devoted to this topic, there seems to be no consensus on the most effective way to extract probabilistic information from SVMs. 
Therefore, the probabilistic nature of KLR makes it preferable to SVM in a number of settings where reliable probabilistic predictions are required to take aware decisions (see e.g., \cite{wiens2019no,kelly2019key,nicora2022evaluating}). Nevertheless, KLR tends to yield denser representations than SVMs, which are natively sparse (only a small subset of data points are needed to specify the decision function) and lead to smaller computational prediction cost. 
}

\textcolor{black}{Sparsity plays a crucial role in machine learning models, as it is closely related to both generalization error and computational efficiency. In statistical learning, the importance of sparsity has been extensively investigated, especially since it leads to sharper generalization bounds (see, e.g., \cite{vandeGeer2008,Bickel2009,Negahban2012}). In kernel machine models like SVMs, the number of support vectors has been shown to strongly affect generalization, with sparse solutions often providing tighter error bounds, as highlighted in the SVM literature \cite{burges1998tutorial,Vapnik2000}. As a consequence, several methods have been proposed to obtain sparse solutions in kernel logistic regression. For instance, \cite{zhu2005kernel,krishnapuram2005sparse,xu2013sparse} show that accurate predictions can be obtained using only a small subset of the data points in the training set. From the computational point of view, sparsity is still of primary importance in practice. Some works, such as \cite{Burges1996,wang2012breaking}, illustrate the benefits of kernel machine models involving a small number of terms (kernel functions) when subject to limited memory and computational time. More recent works address large-scale settings, aiming to reduce RAM usage and CPU time. For instance, \cite{koppel2019parsimonious} studies an online kernel learning framework, \cite{li2022worst} establishes theoretical lower bounds for online kernel selection, and \cite{martin2025scalable} considers Nystr{\"o}m approximations for KLR.}

In this work, we propose an extension of the nonlinear continuous constrained convex formulation adopted in \cite{keerthi2005fast} for training KLR, which induces sparsity in the model without compromising the testing accuracy.
The dual of \textcolor{mygreen}{this} formulation can be adapted to be efficiently solved by a decomposition method. \textcolor{mygreen}{Inspired by the decomposition algorithm in \cite{fan2005working} for SVM}, 
we devise a Sequential Minimal Optimization (SMO)-type training algorithm exploiting second-order information, for which we establish global convergence. 
\textcolor{mygreen}{The proposed formulation and algorithm are} tested on 12 datasets from the literature and compared with Import Vector Machine (IVM) greedy heuristic \cite{zhu2005kernel}, $\ell_{1/2}$-$\text{KLR}$ regularization technique \cite{xu2013sparse}, and SVM. The numerical results shows that \textcolor{mygreen}{our} approach is able to achieve a good trade-off between \textcolor{mygreen}{sparsity and testing accuracy}, while retaining 
the \textcolor{black}{advantages of directly providing estimates of the class 
	probabilities}.

The remainder of the paper is \textcolor{black}{organized} as follows. 
\textcolor{mygreen}{In Section \ref{sec:back}, after recalling the binary KLR model and training formulation, we mention previous work on promoting sparsity in KLR.}
In Section \ref{sec:ocrt}, we describe the proposed nonlinear optimization formulation \textcolor{mygreen}{for training sparse KLR, and its dual}. 
\textcolor{mygreen}{In Section \ref{sec:dec}, we present the SMO-type decomposition method and state the related convergence guarantees which are proved in \ref{subsec:prop_app}.}
The experimental results obtained for 12 datasets from the literature are reported and discussed in Section \ref{sec:experiments}. \textcolor{black}{Finally}, Section \ref{sec:conclusion} \textcolor{black}{contains some} concluding remarks.

\section{\textcolor{black}{Binary KLR and previous works on sparse KLR}}\label{sec:back}

\textcolor{black}{In this section, we introduce the \textcolor{mygreen}{KLR} 
model and the associated primal and dual training formulations. Then, we mention previous work to deal with sparsity in KLR.}

\subsection{\textcolor{black}{Binary KLR model and training formulation}}


In a binary classification problem with $p$ \textcolor{mygreen}{features}, we are given a training set 
$\left \{ (\bm{x}_{i},y_{i})\right \}_{i \in I}$ where $I=\{1,\ldots,N\}$ and each one of the $N$ data points consists of a vector $\bm{x}_{i}\in \mathbb{R}^{p}$ of  \textcolor{mygreen}{features} and an associated label $ y_{i} \in \{-1,1\}$.
Let $\textcolor{black}{\bm{z}} = \varphi(\bm{x})$ \textcolor{black}{denote the higher-dimensional} feature 
space vector 
corresponding to $\bm{x}$, where $\varphi(\cdot)$ represents the feature map which induces the kernel metric $Ker(\bm{x},\hat{\bm{x}})=
\textcolor{black}{\varphi(\bm{x}) \cdot \varphi(\hat{\bm{x}})}$ 
\textcolor{black}{and $\cdot$ denotes the inner product in the $\bm{z}$ space.}  
The binary KLR model can be trained by solving the following unconstrained \textcolor{black}{nonlinear} optimization problem:

\begin{equation}
\label{eq:log_form}
    \min_{\bm{\w},\textcolor{black}{b} } \frac{1}{2} \| \bm{\w}\|^2 + C\sum_{i\in I}g(-y_i(\bm{\w}\cdot\bm{z}_i-b)),
\end{equation}
where \textcolor{black}{$\bm{z}_i = \varphi (\bm{x}_i) $,} $C$ is a regularization parameter, and $g$ is given by:
\begin{equation}
\label{eq:neg_lo_li}
g(\xi) = \text{log}(1+e^\xi).
\end{equation}
\noindent The function $g(\cdot)$ represents the Negative Log-Likelihood (NLL) associated to the conditional probability:
\begin{equation}\label{eq:condprob}
    P(y|\bm{x}) = \frac{1}{1 + e^{-y(\bm{\w} \cdot\varphi(\bm{x})-b)}}.
\end{equation}

\noindent 
\textcolor{black}{By} introducing the auxiliary variables $\xi_i$ with $i \in I$, problem \eqref{eq:log_form} can be rewritten as
\begin{subequations}\label{eq:k-primal}
\begin{equation}\label{eq:k-primal-obj}
\min_{\bm{\w},b,\xi} \quad  
\frac{1}{2}\|\bm{\w}\|^2+C\sum_{i \in I}g(\xi_i)
\end{equation}
\begin{equation}\label{eq:k-primal-con}
\textrm{s.t.} \quad \xi_i= -y_{i}(\bm{\w}\cdot \bm{z}_{i}-b) \quad \forall i \in I . 
\end{equation}
\end{subequations}
\textcolor{black}{As shown in \cite{keerthi2005fast}},
using the Lagrangian function and the optimality conditions, the 
 Wolfe dual of \eqref{eq:k-primal} \textcolor{black}{is as follows} 
\begin{subequations}\label{eq:k-dual}
\begin{equation}\label{eq:k-dual-obj}
\min_{\bm{\alpha}}   \quad  \frac{1}{2} \| \bm{\w}(\bm{\alpha})\|^2 +C\sum_{i\in I}G(\frac{\alpha_i}{C})
\end{equation}
\begin{equation}\label{eq:k-dual-con}
\textrm{s.t.} \quad   \sum_{i \in I} \alpha_i y_i=0, 
\end{equation}
\end{subequations}
where \textcolor{black}{the} $\alpha_i$ with $i \in I$ are the dual variables associated to constraints \eqref{eq:k-primal-con}, 
\textcolor{black}{the} primal and dual variables are linked by the following relations
\begin{equation}\label{eq:pr-du}
    \bm{\w}(\bm{\alpha}) = \sum_{i \in I} \alpha_i y_i \bm{z}_i, \quad \xi_i(\alpha_i) = g'^{-1}(\frac{\alpha_i}{C}),\footnote{Note that 
     $g'(u)=e^u/(1+e^u)$ has \textcolor{black}{its} range in the open interval $(0,1)$ and \textcolor{black}{is invertible on it}.}
\end{equation}
and \textcolor{black}{the} function $G$, defined as
\begin{equation}
    G(\frac{\alpha_i}{C})=\frac{\alpha_i}{C} \xi_i-g(\xi_i),
\end{equation}
is convex. 
Therefore, training a binary KLR model can be formulated as solving the convex \textcolor{black}{constrained nonlinear optimization} problem \eqref{eq:k-dual}, which is very similar to the SVM dual formulation (see e.g., \cite{cortes1995support}). \textcolor{mygreen}{Exploiting this similarity the authors in \cite{keerthi2005fast} devise a SMO-type decomposition algorithm based on first-order information to solve formulation \eqref{eq:k-dual}.}


\subsection{\textcolor{black}{Previous work on sparse KLR}}

In the past twenty years, there have been some attempts to deal with sparsity in KLR. In \cite{krishnapuram2005sparse} the authors combine a bound optimization approach with a coordinate descent update procedure for learning a sparse multi-class classifier within a Bayesian framework. 
In \cite{zhang2012efficient} two conservative online learning algorithms are presented to generate sparse KLR. At each iteration, the current classifier is updated in a stochastic manner using Bernoulli random variables to model whether new data points are considered.

\textcolor{mygreen}{In \cite{zhu2005kernel} the authors propose IVMs where KLR sparsity is promoted by solving the}
problem \eqref{eq:log_form} via a greedy sub-model selection strategy. The approach involves constructing a proxy model for the relation $ \bm{\w}(\bm{\alpha}) = \sum_{i \in I} \alpha_i y_i \bm{z}_i$, by identifying a subset of data points $S$ such that $\sum_{s \in S} \alpha_s y_s \bm{z}_s $ provides a good approximation of $\sum_{i \in I} \alpha_i y_i \bm{z}_i$. 
Since exhaustively searching through all possible subsets of data points is a computationally intractable task\textcolor{black}{, the} authors adopt a forward selection strategy, starting with the empty set $S = \emptyset$ and iteratively adding one data point at a time. The included data point is the one producing the best improvement after a one-step Newton-Raphson method. 
\textcolor{black}{As we shall see in the numerical experiments reported in Section \ref{sec:experiments}, IVMs allow to achieve a significant degree of sparsity. By construction, only a small fraction of data points is considered and then used to index kernel basis functions. However, the greedy heuristic 
tends to cause premature convergence to suboptimal solutions and to compromise accuracy.}

\textcolor{black}{In \cite{roscher2012i2vm,roscher2012incremental} an efficient incremental learning strategy is presented for training sparse IVMs and is applied to the classification of hyperspectral imagery data. New training data points are included to increase the classification accuracy and non-informative data points are deleted to be memory and runtime efficient.
In the comparative experiments of IVM and SVM, the number of import vectors is significantly lower than the number of support vectors, and the probabilities provided by IVM are more reliable than the probabilistic information derived from an SVM’s output using the method described in}
\textcolor{black}{\cite{platt2000}.}

Since regularization is a natural way to promote model sparsity, $\ell_1$, $\ell_2$ and other ad hoc regularization variants have been considered. 
\textcolor{black}{In \cite{xu2013sparse} the $\ell_{1/2}$ regularization is used for kernel logistic regression. In \cite{xu2012l_} a thresholding representation theory for $\ell_{1/2}$ regularization is described and the fast iterative convergent half thresholding algorithm is proposed for a different family of loss functions. A series of experiments on compressed sensing problems show that the convergent half thresholding algorithm is an efficient and effective approach for sparse modeling. In \cite{xu2013sparse} the authors adapt \textcolor{mygreen}{to} KLR the $\ell_{1/2}$-based method and the fast iterative convergent half thresholding algorithm proposed in \cite{xu2012l_}, showing promising results in terms of accuracy and sparsity compared with IVM, two SVM versions and $\ell_1$ regularization.}\par
\blue{The above-mentioned approaches allow to achieve different level of sparsity. Nevertheless, since they are based on heuristic techniques or regularization \textcolor{mygreen}{methods}, it is still a challenging problem to develop a robust training method that yields sparse KLRs with good \textcolor{mygreen}{testing accuracy}.}





\section{A  
	\textcolor{black}{sparsity-inducing approach for training binary KLR}} \label{sec:ocrt}
\textcolor{black}{In this work, aiming at a good trade-off between sparsity and accuracy, we propose an extension of the binary KLR primal formulation \eqref{eq:k-primal} in \cite{keerthi2005fast} specifically designed to promote sparsity in the dual variables, namely, in the vector of the $\alpha_i$ variables, and we derive a bounded regularized (strictly convex) variant of the dual formulation.}

\textcolor{black}{In this section we first describe the sparsity-inducing training formulation and then discuss its main motivation and peculiarities.}

\subsection{\textcolor{black}{A KLR formulation variant to promote sparsity}}

\textcolor{black}{Before 
\textcolor{black}{presenting} the formulation \textcolor{black}{for training sparse KLR}, we briefly recall some key characteristics of the NLL loss used in the KLR model to justify the introduction of \textcolor{black}{a new} variable $\rho$ \textcolor{black}{in the primal formulation}, which plays a crucial role in inducing sparsity.}

\begin{figure}[H]
\centering
\includegraphics[width=0.6\textwidth]{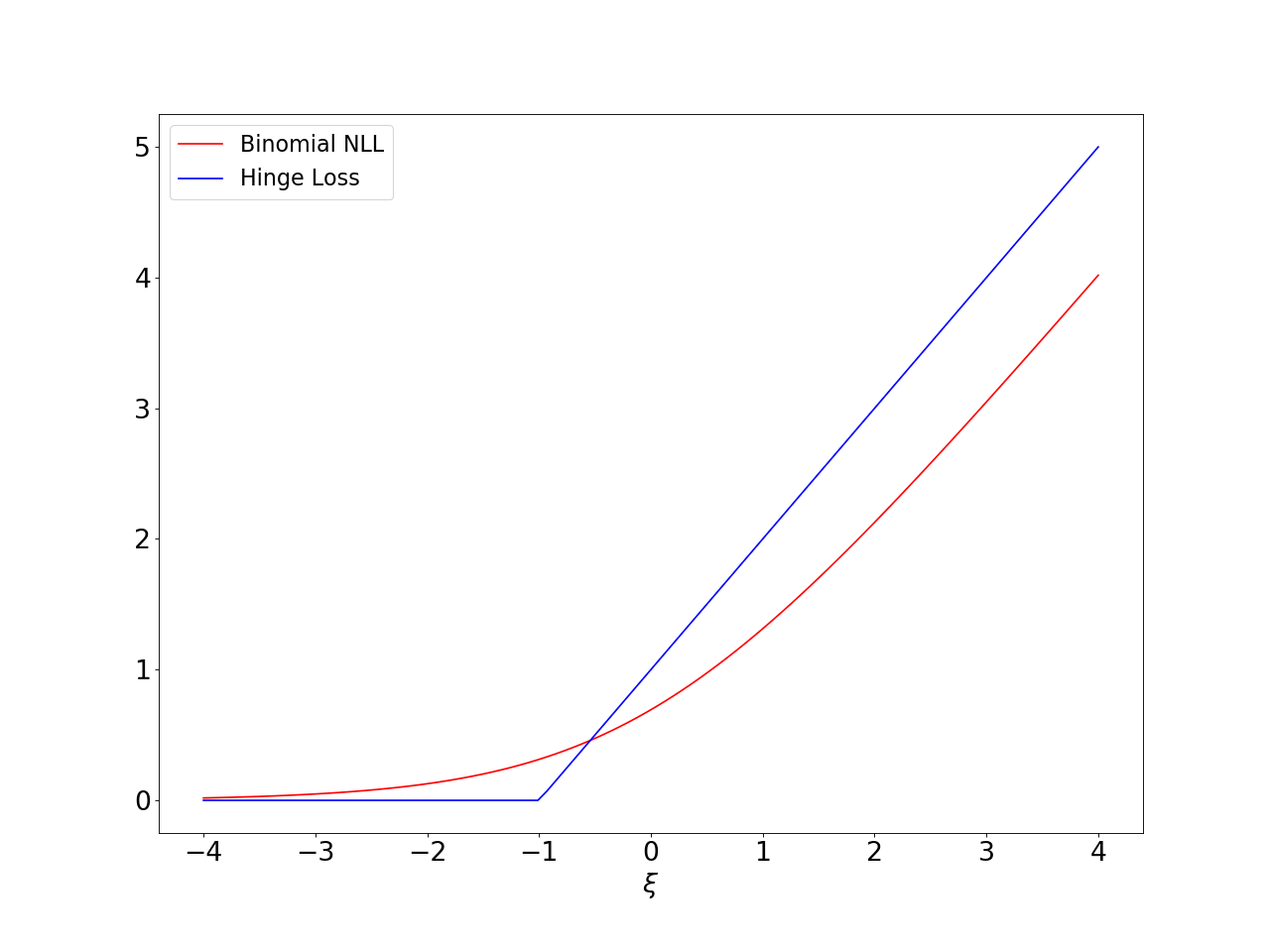}
\caption{\textcolor{black}{Plots} of the Hinge loss used
in Support Vector Machines
\textcolor{black}{in blue and of} the Binomial Negative Log-Likelihood \textcolor{black}{used in} logistic regression \textcolor{black}{in} red.}\label{fig:losses}
\end{figure}

\textcolor{black}{In the literature, several authors \cite{zhu2005kernel,keerthi2005fast} have analysed the similarity between the NLL and the hinge loss, which is often applied in ML models such as SVM. As shown in Figure \ref{fig:losses}, the NLL function approximates the hinge loss. However, whereas the hinge loss is capable of inducing sparsity due to its truncation property, the NLL function lacks this capability. As a result, the KLR model typically yields an optimal solution where $\alpha_i > 0$ for all $i \in I$. From the Representer Theorem \cite{scholkopf2001generalized}, it is known that the classifier parameters $\bm{\w}$ of the KLR can be expressed as a weighted sum of the data points mapped into the higher-dimensional feature space, where the weights correspond to the variables $\alpha_i$. The aim is to modify formulation \eqref{eq:k-primal} in order to reduce the impact in the objective function of the NLL terms corresponding \textcolor{black}{to the easy-to-classify data points}, 
i.e., those with sensibly negative values of the associated $\xi_i$ variable. For this purpose, we define a nonnegative slack variable $\rho$ that adjusts the $\xi_i$ values corresponding to the \textcolor{black}{easy-to-classify} data points, 
 reducing their associated dual variable values $\alpha_i$.}



\begin{figure}[h]
\centering
\begin{subfigure}{0.48\textwidth}
    \includegraphics[width=\textwidth]{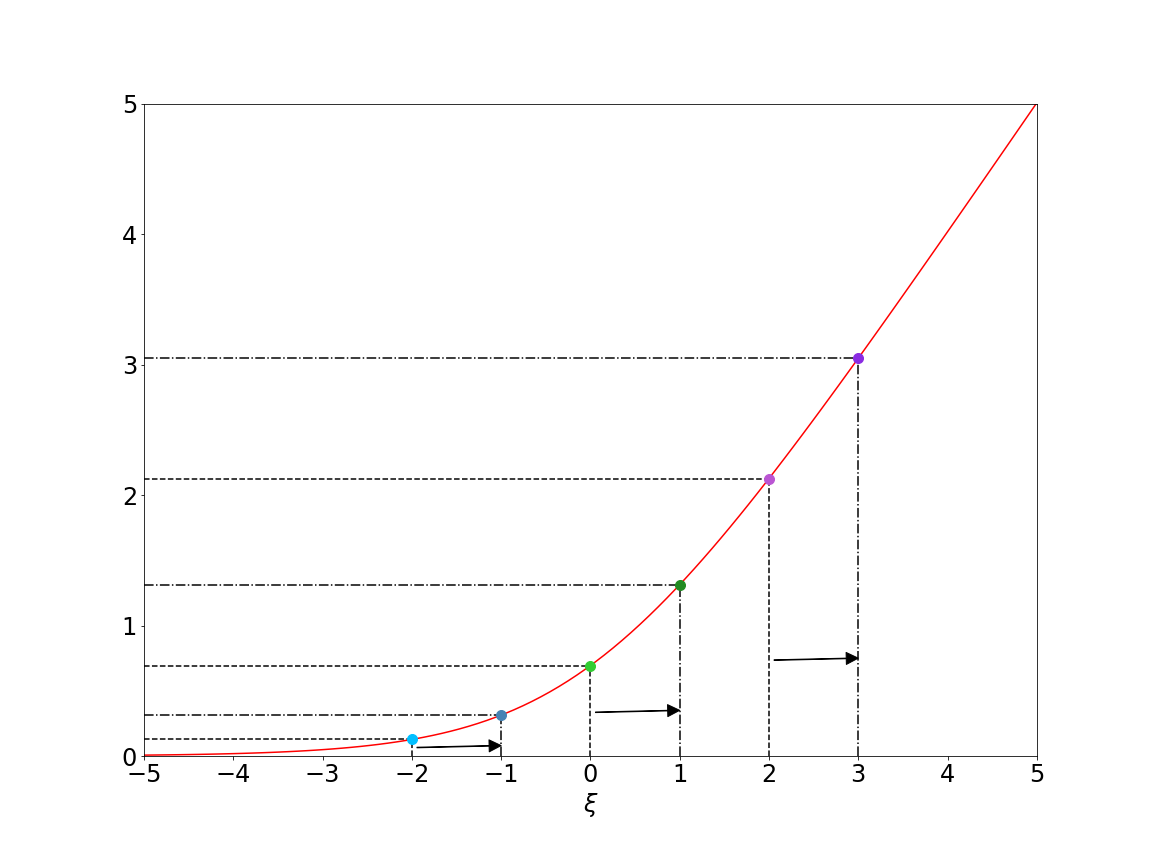}
    \caption{Effect of a right-shift of a $\xi$ variable on the NLL function for different $\xi$ levels.}
    \vspace{9.4pt}
    \label{fig:first}
\end{subfigure}
\hfill
\begin{subfigure}{0.48\textwidth}
    \includegraphics[width=\textwidth]{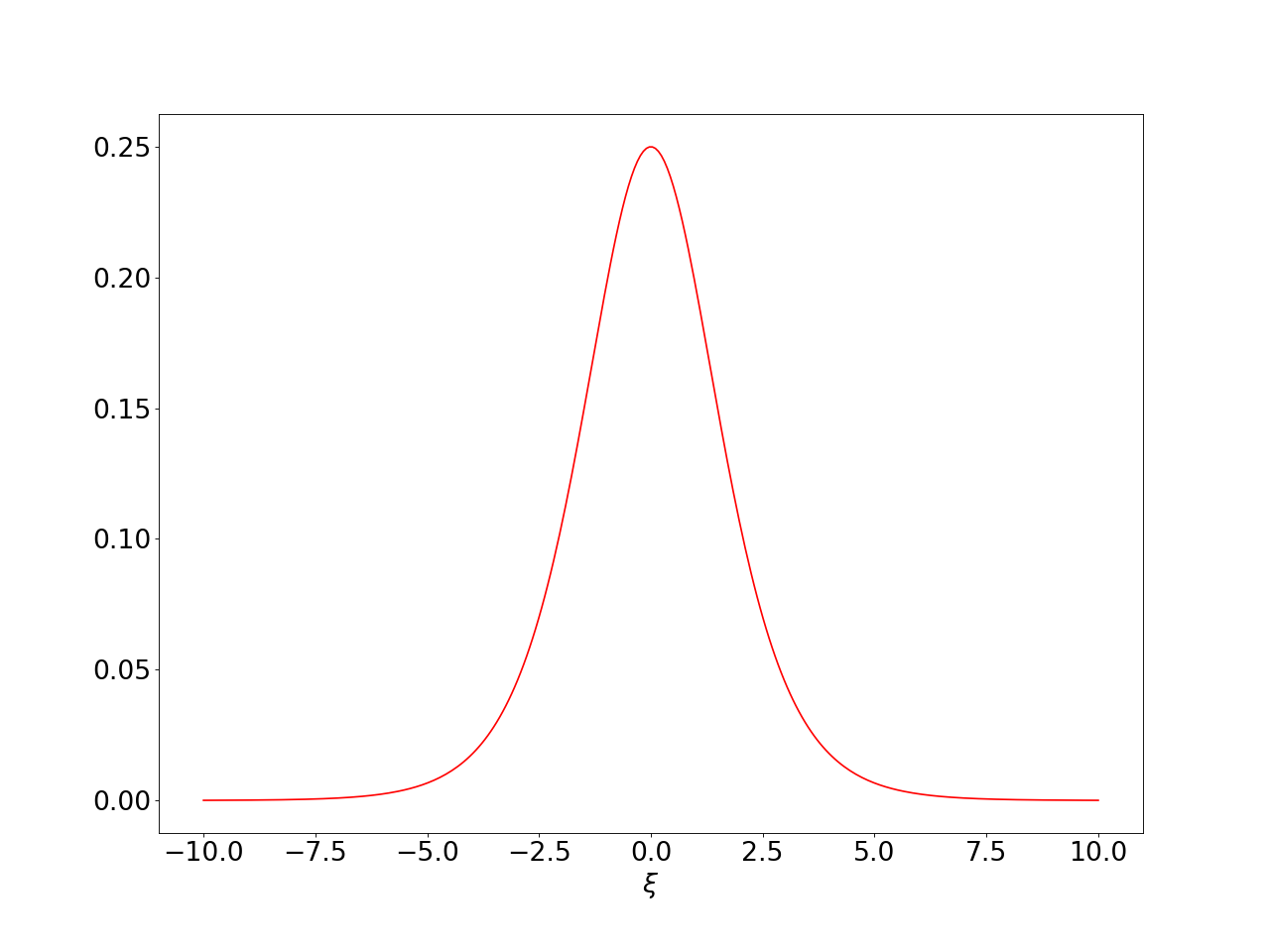}
    \caption{Second derivative of the NLL function. $\xi$ values away from zero have a flat profile, while around zero there is a positive peak.}
    \label{fig:binomial_deriv}
\end{subfigure}
\caption{\textcolor{black}{Illustration of the negative log-likelihood function and its second derivative with respect to the parameter $\xi$.}}

\end{figure}

\textcolor{black}{The 
formulation that we propose for training sparse KLR} is as follows:
\begin{subequations}\label{eq:sparse-primal}
    \begin{equation}\label{eq:form_cons}
    \min_{\bm{\w}, \textcolor{black}{b} ,\bm{\rho}}   \quad  \frac{1}{2} \| \bm{\w}\|^2 + C\sum_{\textcolor{black}{i \in I}}g(\xi_i) - \nu \rho  
    \end{equation} 
    \begin{equation}\label{eq:sparse-primal-con}
    s.t. \quad  \quad  \xi_i -\rho = -y_i (\bm{\w}\cdot\bm{z}_i-b)  \quad  \quad    
    \textcolor{black}{i \in I}
    \end{equation}
    \begin{equation}\label{eq:sparse-prima-non}
     \rho \geq 0. 
    \end{equation}
\end{subequations}
\textcolor{black}{The constraints \eqref{eq:sparse-primal-con}} expressing the decision for each data point are modified with respect to the original formulation \eqref{eq:k-primal} by \textcolor{black}{subtracting} a nonnegative slack variable $\rho$ to the left-hand-side of \textcolor{black}{all} the equality \textcolor{black}{constraints \eqref{eq:k-primal-con}}. 
This implies a general increment of variables $\xi_i$ for the constraints to be satisfied. Notice that a right shifting of variable $\xi_i$ is equivalent to a left ``translation" of the corresponding NLL function. As depicted in Figure \ref{fig:first}, the consequence is that the NLL errors of data points with $\xi_i$ values around zero (which are strongly relevant in determining the final classifier), would significantly increase. Conversely, errors of data points with sharply negative $\xi_i$ values lying in the flat left region of the NLL function, whose classification is trivial, would not change significantly. 

This effect can also be viewed in terms of derivatives of the NLL function. Indeed, the profile of the NLL second derivatives, shown in Figure \ref{fig:binomial_deriv}, has a remarkable positive peak around zero, and it is 
\textcolor{black}{almost flat and close to zero sufficiently far from it}. Therefore, an increase 
\textcolor{black}{in the value of} $\xi_i$ produces a relatively larger first derivatives increase for values \textcolor{black}{of $\xi$} around zero, than for 
\textcolor{black}{strongly} negative/positive values. This fact is expected to strongly influence the training optimization phase, as it leverages first derivatives information.\par
To amplify the effect of the variable $\rho$, the latter is maximized by adding \textcolor{black}{the} term $- \nu \rho $ in the objective function \eqref{eq:sparse-prima-non}, with $\nu > 0$. However, $\rho$ cannot grow arbitrarily as, due to constraints \eqref{eq:sparse-primal-con}, too large values of $\rho$ would imply too large values of the NLL error terms $C\sum\limits_{\textcolor{black}{i \in I}}g(\xi_i)$. The desired trade-off between maximization of $\rho$ and minimization of the NLL errors can be calibrated by \textcolor{black}{selecting the values of the $C$ and $\nu$ parameters}.\par 

By considering the Lagrangian function of \eqref{eq:sparse-primal} 
\begin{equation}
    \label{eq:lag_f}
   \textcolor{black}{L(\bm{\w},b,\bm{\xi},\rho)} = \frac{1}{2} \| \bm{\w}\|^2 + C\sum_{\textcolor{black}{i \in I}}g(\xi_i) - \nu \rho +\sum_{\textcolor{black}{i \in I}} \alpha_i[-\xi_i +\rho -y_i (\bm{\w}\cdot\bm{z}_i-b)] - 
    \textcolor{black}{\beta} \rho 
\end{equation}
with $\textcolor{black}{\beta} \geq 0$, the optimality conditions are
\begin{align}
    & \nabla_{\bm{\w}}L = \bm{\w} - \sum_{\textcolor{black}{i \in I}} \alpha_i y_i \bm{z}_i = \textcolor{black}{\bm{0}} \label{eq:oc_w}\\
    & \frac{\partial L}{\partial b} = \sum_{\textcolor{black}{i \in I}} \alpha_i y_i = 0  \label{eq:oc_b} \\
    & \frac{\partial L}{\partial \xi_i} = C g'(\xi_i) - \alpha_i = 0 \quad  \quad
    \textcolor{black}{i \in I} \label{eq:oc_xi}  
    \\
    & \frac{\partial L}{\partial \rho} = -\nu + \sum_{\textcolor{black}{i \in I}} \alpha_i - \textcolor{black}{\beta}  
    = 0. \label{eq:oc_rho}
\end{align}

Thanks to \eqref{eq:oc_w} and \eqref{eq:oc_xi}, we can write $\bm{\w}$ and $\xi_i$ as functions of the variables $\alpha_i$:
\begin{equation}
    \bm{\w}(\bm{\alpha}) = \sum_{\textcolor{black}{i \in I}} \alpha_i y_i \bm{z}_i, \quad \xi_i(\alpha_i) = g'^{-1}(\frac{\alpha_i}{C}). \label{eq:w_xi}
\end{equation}

As shown in \cite{keerthi2005fast}, it is possible to define a function
\begin{equation}
\label{eq:g_func}
    G(\delta) = \delta\xi_i - g(\xi_i), \quad \text{with} \; \delta = \frac{\alpha_i}{C}
\end{equation}
whose derivative expresses the inverse function $g'^{-1}$ in \eqref{eq:w_xi}:
$$\frac{d G}{d \delta} = \delta\frac{d \xi_i}{d \delta} + \xi_i - g'(\xi_i)\frac{d \xi_i}{d \delta} = \xi_i = g'^{-1}(\delta).$$
For the case of logistic regression where $g$ is expressed as \eqref{eq:neg_lo_li} we have:
\begin{align}\label{eq:gG}
    & g'^{-1}(u) = \text{log}(\frac{u}{1-u})
    & G(\delta) = \delta \text{log}(\delta) + (1 -\delta) \text{log}(1-\delta).
\end{align}

Using equations \textcolor{black}{\eqref{eq:w_xi} and \eqref{eq:g_func}} and \textcolor{black}{recalling that $\beta \geq 0$}, 
the Wolfe dual (which corresponds to the maximization of $L(\bm{\w},b,\bm{\xi},\rho)$ subject to \eqref{eq:oc_w}-\eqref{eq:oc_rho}) 
 can be rewritten as:
\begin{subequations}\label{eq:final_p}
    \begin{equation}\label{eq:final_f} 
    \min_{\bm{\alpha}}   \quad  \frac{1}{2} \| \bm{\w}(\bm{\alpha})\|^2 +C\sum_{\textcolor{black}{i \in I}} G(\frac{\alpha_i}{C})
    \end{equation}
    \begin{equation}\label{eq:final_c1} 
         s.t. \quad    \sum_{\textcolor{black}{i \in I}} \alpha_i y_i=0 
    \end{equation}
   \begin{equation}\label{eq:final_c2}
    \quad  \quad   -\nu \geq  -\sum_{\textcolor{black}{i \in I}} \alpha_i.
    \end{equation}
\end{subequations}
Instead of directly solving formulation \eqref{eq:final_p} which includes the complicating constraint \eqref{eq:final_c2}, we consider the following ``regularized" \textcolor{black}{formulation}
\begin{align}\label{eq:regularized}
    \min_{\bm{\alpha}} &  \quad  \frac{1}{2} \| \bm{\w}(\bm{\alpha})\|^2 +C\sum_{i\in I}G(\frac{\alpha_i}{C})- \lambda \sum_{i \in I} \alpha_i \\
    s.t. \quad & \quad  \sum_{i \in I} \alpha_i y_i=0, \nonumber
\end{align}
where constraint \eqref{eq:final_c2} is replaced by a sort of regularization term $- \lambda \sum_i \alpha_i$ in the objective function, where the weight $\lambda \geq 0$ is a hyperparameter to be tuned. Notice that 
the hyperparameter $\nu$ in constraint \eqref{eq:final_c2} is taken into account by the hyperparameter $\lambda$ in the objective function.\par

\textcolor{black}{Some simple observations are in order concerning the impact of the $\rho$ variable introduced in formulation \eqref{eq:sparse-primal} with respect to the other primal variables $\xi_i$ with $i \in I$.
The optimal solution $(\hat{\bm\omega},\hat b, \hat{\bm\xi})$ of \eqref{eq:sparse-primal} with $\nu=0$ (which is equivalent to the optimal solution of the original formulation \eqref{eq:k-primal}) 
is the best one (ideal) in terms of trade-off between $\frac{1}{2} || \bm{\w}||^2$ and $\sum_{{i \in I}}g(\xi_i)$ for any given value of $C$, but it is generally not sparse. 
Let us consider a sufficiently small positive value of $\nu$ which leads to a small positive optimal $\rho$ value, and to an optimal solution of \eqref{eq:sparse-primal} that is a ``local" perturbation of $(\hat{\bm\omega},\hat b, \hat{\bm\xi})$. 
This perturbation implies that constraint \eqref{eq:k-primal-con} is ``replaced" with constraint \eqref{eq:sparse-primal-con}, and hence that the ideal but not sparse 
triplet $(\hat{\bm\omega},\hat b, \hat{\bm\xi})$ must be modified to be feasible with respect to \eqref{eq:sparse-primal-con}. 
Since for sufficiently small values of $\mu$ the optimal solution of \eqref{eq:sparse-primal} is not expected to deviate too much from $(\hat{\bm\omega},\hat b, \hat{\bm\xi})$, a possible way to enforce feasibility, while preserving the decision boundary $(\hat{\bm\omega},\hat b)$, is to uniformly increase all $\hat\xi_i$ values by $\rho$. However, since the increase of each $\xi_i$ has a different impact on the corresponding error terms, a (slight) modification of $(\hat{\bm\omega},\hat b)$ in the right-hand side of \eqref{eq:sparse-primal-con} would imply heterogeneous updates of the variables $\xi_i$ which could be preferable in terms of the objective function value. In such a case, some $\xi_i$ variables would increase and some others would decrease, which, given the monotonic connection between primal variables $\xi$ and dual variables $\alpha_i$ (see \eqref{eq:oc_xi}), tend to promote dual sparsity. As empirically shown in Section \ref{subsec:c-lambda}, for reasonable values of the primal (dual) hyperparameter $\nu$ ($\lambda$) the norm of the decision boundary hyperplane $(\hat{\bm\omega},\hat b)$ increases, while its position and orientation are not significantly affected, which preserves the same classification rates as the ones obtained by the ideal triplet  $(\hat{\bm\omega},\hat b, \hat{\bm\xi})$. This is consistent with the fact that positive values of $\rho$ tend to increase the values of the variables $\xi_i$, and consequently of the relative weight of the error term $C\sum_{{i \in I}}g(\xi_i)$ in the objective function. This, combined with the presence of the new term $-\nu \rho$ in formulation \eqref{eq:sparse-primal}, reduces the relative weight of the regularization term $\frac{1}{2} \| \bm{\w}(\bm{\alpha})\|^2$.}

\subsection{\textcolor{black}{Some properties}}\label{sec:useful_prop}
\textcolor{black}{In this subsection we describe some properties of the optimal dual solution $\bm{\alpha}$ and the connection between the regularization hyperparameter $\lambda$ and the primal variable $\rho$ of formulation \eqref{eq:sparse-primal}. Additional properties of $\bm{\alpha}$ as $\lambda$ tends to infinity can be found in \ref{sec:bounds_lambda}.} 




\textcolor{black}{
Since the function $G(\cdot)$ is a entropy function, every dual variable $\alpha_i$ clearly takes values within the compact interval $[0,C]$. 
For every $i \in I$ and any given $\lambda$, let $\bm{\alpha}(\lambda)$ denote the corresponding optimal solution of \eqref{eq:regularized}.}

\textcolor{black}{
In \ref{sec:bounds_lambda},  we prove that for any value of $\lambda \in \mathbb{R}$ all the components $\alpha_i(\lambda)$ of the optimal solution never attain the above-mentioned bounds.
}

\textcolor{black}{\begin{lemma}\label{lem:unbound_alpha}
For any value of $\lambda \in \mathbb{R}$ the optimal solution of formulation \eqref{eq:regularized} $\bm{\alpha}(\lambda)$ is such that $$0 < \alpha_i(\lambda) < C \quad \forall i \in I.$$
\end{lemma}}

\textcolor{black}{Now we analyze the connection between the primal variable $\rho$ and the dual formulation \eqref{eq:regularized}. For each $\lambda \in \mathbb{R}$, the corresponding optimality conditions associated with the dual problem \eqref{eq:regularized} lead (via substitution) to the following equations:}

\textcolor{black}{\begin{equation} \label{eq: prof_bound2}
   \lambda - y_i \sum\limits_{j \in I}\alpha_j y_j K_{ij} + y_i \psi = g'^{-1} (\frac{\alpha_i}{C}) \;\; \qquad \forall i \in I,
\end{equation}}

\noindent \textcolor{black}{where $\psi$ is the Lagrangian multiplier associated to the equality constraint of \eqref{eq:regularized}.}

\textcolor{black}{Moreover, for every optimal solution of the dual problem \eqref{eq:regularized} the optimality condition of the primal problem \eqref{eq:oc_xi} and \eqref{eq:sparse-primal-con} are also satisfied. Therefore:}

\textcolor{black}{$$\rho - y_i (\bm{\omega}\cdot\bm{z}_i - b) = g'^{-1} (\frac{\alpha_i}{C}) \;\; \qquad \forall i \in I,$$}

\noindent\textcolor{black}{which, given \eqref{eq:oc_w}, can be rewritten as:
\begin{equation} \label{eq: prof_bound1}
    \rho - y_i \sum\limits_{j \in I}\alpha_j y_j K_{ij} + y_i b = g'^{-1} (\frac{\alpha_i}{C}) \;\; \qquad \forall i \in I.
\end{equation}}

\textcolor{black}{Any optimal solution $\bm{\alpha}(\lambda)$ of the dual formulation \eqref{eq:regularized} must satisfy the system of equations \eqref{eq: prof_bound1} in the variables $\rho$ and $b$, as well as the system of equations \eqref{eq: prof_bound2} in the variables $\lambda$ and $\psi$. Since the number  $N$ of equations is larger than the number of variables ($2$), both linear systems admit a unique solution, and these solutions coincide. This directly leads to the following result.}

\begin{observation}
    \textcolor{black}{For each $\lambda \in \mathbb{R}$ in the dual fromulation \eqref{eq:regularized}, the associated primal optimal solution, denoted as $(\bm\omega(\lambda), b(\lambda),\rho(\lambda))$ is such that $\rho(\lambda) = \lambda$. Hence, the optimal value of the primal variable $\rho$ 
corresponds exactly to the value of the coefficient of the regularization term in the dual formulation \eqref{eq:regularized}.}
\end{observation}

\textcolor{black}{From the above observation, we derive an upper bound on the values of $\lambda$ which can be used in practice. Since values of $G(\cdot)$ cannot be reliably computed when $\alpha_i$ approaches $0$ or $C$, thanks to Lemma \ref{lem:unbound_alpha}, we consider a modified version of formulation \eqref{eq:regularized}. Given a tolerance $\gamma > 0$, we define $\overline{0} = \gamma$ and $\overline{C} = C - \gamma$ and add the auxiliary constraints $\overline{0} \leq \alpha_i \leq \overline{C}$ for all $i \in I$ in formulation  \eqref{eq:regularized}. This is similar to what was done to avoid numerical issues in \cite{keerthi2005fast}, where the auxiliary bounds are not explicitly added to the dual of the original formulation \eqref{eq:k-primal} but are implicitly taken into account in the training algorithm.}

\textcolor{black}{
Let $I_+=\{i \in I:y_i=+1\}$, $I_-=\{i\in I:y_i=-1\}$, $n_{+}=|I_{+}|$, $n_{-}=|I_{-}|$ and the minority class be the one with the smallest number of data points. Assume without loss of generality that $I_-$ is the minority class.
Given any tolerance $\gamma > 0$, it is possible to obtain an upper bound on the value of $\lambda$ beyond which the optimal solution $\bm{\alpha}(\lambda)$ satisfies $\alpha_i(\lambda) \geq C -\gamma$ for every data point $(\bm{x}_i,y_i)$ in the minority class with $i \in I$. Therefore, since $\bm{\alpha}_i(\lambda) \gg 0$ for $i \in I_{-}$, the optimal solution is not sparse at least over the components of $\bm{\alpha}$ corresponding to the minority class.}

\textcolor{black}{
In particular, in \ref{sec:bounds_lambda} we prove the following result.}

\textcolor{black}{\begin{proposition}\label{prop:final_bound}
Assume that $I_-$ is the minority class. Let $\bm\alpha(0)$ denote the optimal solution of formulation \eqref{eq:regularized} with $\lambda = 0$. Consider any positive tolerance $\gamma$ such that $\gamma< C-\max\limits_{i \in I_{-}}\alpha_i(0)$, and define $\bar{C} = C - \gamma$. For any
\begin{equation}\label{eq:ub}
    \lambda \geq \max\limits_{i \in I_-} \sum\limits_{j \in I} (\bar{C}-\alpha_j(0)) y_j K_{ji},
\end{equation}
the optimal solution ${\bm\alpha}(\lambda)$ of formulation \eqref{eq:regularized} satisfies $\alpha_i(\lambda) \geq \bar{C}$ for all $i \in I_-$. 
\end{proposition}}
\textcolor{black}{Notice that in case of a perfectly balanced dataset (i.e. $n_+=n_-$), the result of Proposition \ref{prop:final_bound} is valid for every data point of the training set. Indeed, in the case of a balanced dataset, considering $\gamma< C-\max\limits_{i \in I}\alpha_i(0)$, for any $\lambda \geq \max\limits_{i \in I} \sum\limits_{j \in I} (\bar{C}-\alpha_j(0)) y_j K_{ji}$, we have that $\alpha_i(\lambda) \geq \bar{C}$ for all $i \in I$.}


\textcolor{black}{From a practical point of view Proposition \ref{prop:final_bound} can be exploited to restrict the search space for the value of the hyperparameter $\lambda$, as we aim at a sparse dual variable vector $\bm\alpha(\lambda)$. For datasets with imbalanced classes, the inequality \eqref{eq:ub} provides an upper bound for $\lambda$ to avoid that all $\alpha_i(\lambda)$ for all $i \in I_{-}$ exceed the value $\bar C$, preventing sparsity in the minority class. While, for perfectly balanced datasets ($n_+=n_-$), the same inequality can be used to avoid that all $\bm\alpha(\lambda)$ components exceed the value $\bar C$, preventing sparsity over all the components of $\bm\alpha$. Even when the imbalance is mild and the minority class represents slightly less than $50\%$  of data points ($\frac{n_-}{N} <0.5$), selecting a value of $\lambda$ above the threshold in \eqref{eq:ub} implies that the proportion of the nonzero $\bm\alpha(\lambda)$ components is at least $\frac{n_-}{N}$.}

\subsection{\textcolor{black}{A bounded formulation version}}

Therefore, \textcolor{black}{for practical purposes we propose to train the following bounded} regularized version of formulation \eqref{eq:final_p}:
\begin{subequations}\label{eq:2BM}
    \begin{equation}
                \min_{\bm{\alpha}}  \quad  f(\bm{\alpha}) = \frac{1}{2} \| \bm{\w}(\bm{\alpha})\|^2 +C\sum_{i\in I}G(\frac{\alpha_i}{C})- \lambda \sum_{i\in I} \alpha_i\label{eq:final} 
    \end{equation}
    \begin{equation}
                s.t. \;\;  \quad  \sum_{i\in I} \alpha_i y_i=0 \qquad \qquad \qquad \quad\label{eq:final_c}
    \end{equation}
    \begin{equation}
                \quad  \quad  \overline{0} \leq \alpha_i \leq \overline{C} \qquad \forall i \in I. \label{eq:final_compact}
\end{equation}

\end{subequations}
\noindent Notice that the objective function \eqref{eq:final} can be expressed as a function of only \textcolor{black}{the $\alpha_i$} variables as
\begin{equation}
    \label{eq:final_alpha}
    \frac{1}{2} \sum_{i\in I} \sum_{j\in I} y_i y_j \alpha_i \alpha_j K_{ij}  +C\sum_{i\in I}G(\frac{\alpha_i}{C})- \lambda \sum_{i\in I} \alpha_i,
\end{equation}
where $K_{ij} = Ker(\bm{x}_i,\bm{x}_j)$ is the $ij$ element of the $N \times N$ kernel matrix $K$.

\textcolor{black}{According the Karush-Kuhn-Tucker (KKT) conditions,} $\bm{\alpha}$ is a stationary point of \textcolor{black}{\eqref{eq:final}-\eqref{eq:final_compact}}
if and only if there exist two nonnegative vectors $\bm{\sigma}$, $\bm{\eta}$ and a scalar $b$ such that 
$$ \nabla f(\bm{\alpha})+b\bm{y} = \bm{\eta} - \bm{\sigma},$$
$$\eta_i(\alpha_i-\overline{0}) = 0,\, \sigma_i(\overline{C} - \alpha_i)=0,\, \eta_i \geq 0,\, \sigma_i \geq 0\; \; \quad \forall i \in I.$$

The KKT conditions \textcolor{black}{can be rewritten} as
\begin{align}
    \nabla f(\bm{\alpha})_i + b y_i \geq 0 \qquad \text{if}\; \alpha_i < \overline{C},\label{eq:cond1}\\
    \nabla f(\bm{\alpha})_i + b y_i \leq 0 \qquad \text{if}\; \alpha_i > \overline{0}.\label{eq:cond2}
\end{align}

Defining as in \cite{chen2006study}
\textcolor{black}{
\begin{align}
\label{eqn:I_up_low}
\begin{split}
I_{\text{up}}(\boldsymbol{\alpha}) = \{i\,|\, \alpha_i < \overline{C},y_i = 1 \;\text{or}\; \alpha_i >\overline{0}, y_i = -1 \},    \\
I_{\text{low}}(\boldsymbol{\alpha}) = \{i\,|\, \alpha_i < \overline{C},y_i = -1 \;\text{or}\; \alpha_i >\overline{0}, y_i = 1 \}
\end{split}
\end{align}
}
the conditions \eqref{eq:cond1} \textcolor{black}{and} \eqref{eq:cond2} are equivalent to
\begin{align*}
     -y_i \nabla f(\bm{\alpha})_i \leq b \qquad &\forall\, i \in \, I_{\text{up}}(\bm{\alpha}),\\
     -y_i \nabla f(\bm{\alpha})_i \geq b \qquad &\forall\, i \in \,  I_{\text{low}}(\bm{\alpha}).
\end{align*}
Thus a \textcolor{black}{feasible} solution $\bm{\alpha}$ is a stationary point of \textcolor{black}{\eqref{eq:final}-\eqref{eq:final_compact}} if 
\begin{equation}
    \max_{i \in I_{\text{up}}(\bm{\alpha})} -y_i \nabla f(\bm{\alpha})_i \leq \min_{i \in I_{\text{low}}(\bm{\alpha})} -y_i \nabla f(\bm{\alpha})_i.\label{eq:opt_cond}
\end{equation}

\subsection{Motivation for the sparsity inducing formulation}\label{subsec:motivating}

In this section, we provide illustrative examples to analyze the way formulation \eqref{eq:sparse-primal} induces sparsity in the 
KLR model. 
\textcolor{black}{In particular, we show} how the decision boundary, the sparsity level, and the slack variable $\rho$ 
\textcolor{black}{vary for} different values of the hyperparameter $\nu$, and
\textcolor{black}{we point out the differences with respect to heuristics based on thresholding}  
the dual variables $\alpha_i$. For the ease of representation, we consider a linear kernel and a \blue{synthetic not linearly separable dataset, referred to as Synth, for which reasonable linear decision boundaries (hyperplanes) exist. Synth includes $p=2$ features and consists \textcolor{black}{of} 99 data points (50 belonging to the first class and 49 to the second one).}

\subsubsection{The role of the $\rho$ variable in promoting sparsity}\label{sec:role_rho}

\blue{By comparing the original formulation \eqref{eq:k-primal} and} \textcolor{black}{the proposed \textcolor{black}{sparse KLR} formulation \eqref{eq:sparse-primal}}, \blue{for any identical value of hyperparameter $C$ and sufficiently large values of hyperparameter $\nu$, we show that the presence of variable $\rho$ in \eqref{eq:sparse-primal} leads to a different decision boundary  $(\bm \w,  b)$ with respect to the one obtained with the formulation \eqref{eq:k-primal} (formulation \eqref{eq:sparse-primal} with $\nu=0$) which is determined by a relatively smaller subset of the training data points. As we will see via some illustrative examples, this behavior tends to promote sparsity in the associated dual variables $\alpha_i$.}\par

\blue{First, we observe that for any given value of $C$ there exists a $\bar \nu>0$ such that any given $\nu >\bar \nu$ implies that the optimal value of $\rho$ is strictly positive. Since problems \eqref{eq:log_form} and \eqref{eq:sparse-primal} are \textcolor{black}{strictly} convex, we denote by $(\hat{\bm{\w}},\hat{b})$ and $(\bm{\w}^*,b^*,\rho^*)$ the unique optimal solutions of \eqref{eq:log_form} and \eqref{eq:sparse-primal}, respectively.}
\blue{For the sake of computational simplicity, let us consider for problem \eqref{eq:sparse-primal} the following equivalent bound constrained formulation obtained by substituting in the objective function the $\xi_i$ variables according to the equality constraints \eqref{eq:sparse-primal-con} :} 

\begin{subequations}\label{eq:sparse-primal2}
    \begin{equation*}
    \min_{\bm{\w}, \textcolor{black}{b} ,\bm{\rho}}  \bar{f}(\bm{\w}, \textcolor{black}{b} ,\rho)=  \frac{1}{2} \| \bm{\w}\|^2 + C\sum_{\textcolor{black}{i \in I}}g(\rho -y_i (\bm{\w}\cdot\bm{z}_i-b)) - \nu \rho  
    \end{equation*} \begin{equation*}\label{eq:sparse-prima-non_proof}
     \rho \geq 0. 
    \end{equation*}
\end{subequations}

\noindent 
\textcolor{black}{To guarantee that $\rho^*>0$, the partial derivative of $\bar{f}$ with respect to $\rho$ at $(\bm{\w}^*,b^*,0)$ should be negative, i.e. $\frac{\partial \bar{f}(\bm{\w}^*,b^*,0)}{\partial \rho} < 0 $, otherwise $(\bm{\w}^*,b^*,0)$ is optimal due to the constraint $\rho \geq 0$.} Notice that in such a case $(\bm{\w}^*,b^*)$ coincides with $(\hat{\bm{\w}},\hat{b})$. Therefore, to obtain $\rho^*>0$ we need to impose that

$$\frac{\partial \bar{f}(\bm{\w}^*,b^*,0)}{\partial \rho} = \frac{\partial \bar{f}(\hat{\bm{\w}},\hat{b},0)}{\partial \rho} = -\nu + C \sum_{i \in I}\frac{1}{1+e^{y_i(\hat{\bm{\w}}\cdot\bm{z}_i - \hat{b})}}< 0,$$
\noindent implying that

\begin{equation}\label{eq:nu_big}
\nu > \bar{\nu} =  C \sum_{i \in I}\frac{1}{1+e^{y_i(\hat{\bm{\w}}\cdot\bm{z}_i - \hat{b})}}.
\end{equation}

\blue{Given any fixed values of $C$ and $\nu > \bar{\nu}$, for any $(\bm{\w},b) \in \mathbb{R}^{p+1}$ the positivity of the corresponding optimal $\rho$ together with constraint \eqref{eq:sparse-prima-non} lead to a general increase in the variables $\xi_i$ with respect to formulation \eqref{eq:k-primal}. This results in an increase in the weight of the second term $ \sum_{i \in I}g(\xi_i)$ of the objective function \eqref{eq:form_cons}.}
While all the variables $\xi_i$ increase by the same amount $\rho$, the increase in each $g(\xi_i)$ term of the objective function substantially differs depending on the data point. Clearly, given the negative log-likelihood $g$ depicted in Figure \ref{fig:first}, the data points with larger $\xi_i$ (i.e., those closer to the decision boundary, whether correctly or incorrectly classified) will have a greater impact on the increase in the error term. We \textcolor{black}{remark} that,
due to the \blue{nature of $g$, this is not equivalent to increasing $C$ since this would increment all $g(\xi_i)$ terms by the same multiplicative factor for all the data points.}

\blue{The general increase in $ \sum_{i \in I}g(\xi_i)$ leads to a variation of the components of the decision boundary vector $\bm{\w}$, as $\|\bm{\w}\|$ tends to be larger due to a reduced impact of the first term $\frac{1}{2}\|\bm{\w}\|^2$ in the objective function. 
Since the increase in $ \sum_{i \in I}g(\xi_i)$ is more affected by the increment in the terms $g(\xi_i)$ related to the data points which are closer to the decision boundary (those with larger $\alpha_i$ values), the change in the components of $\bm{\w}$ is more influenced by such data points.} 



\begin{figure}[htbp]
    \centering
    \begin{minipage}[b]{0.48\textwidth}
        \centering
        \includegraphics[width=\textwidth]{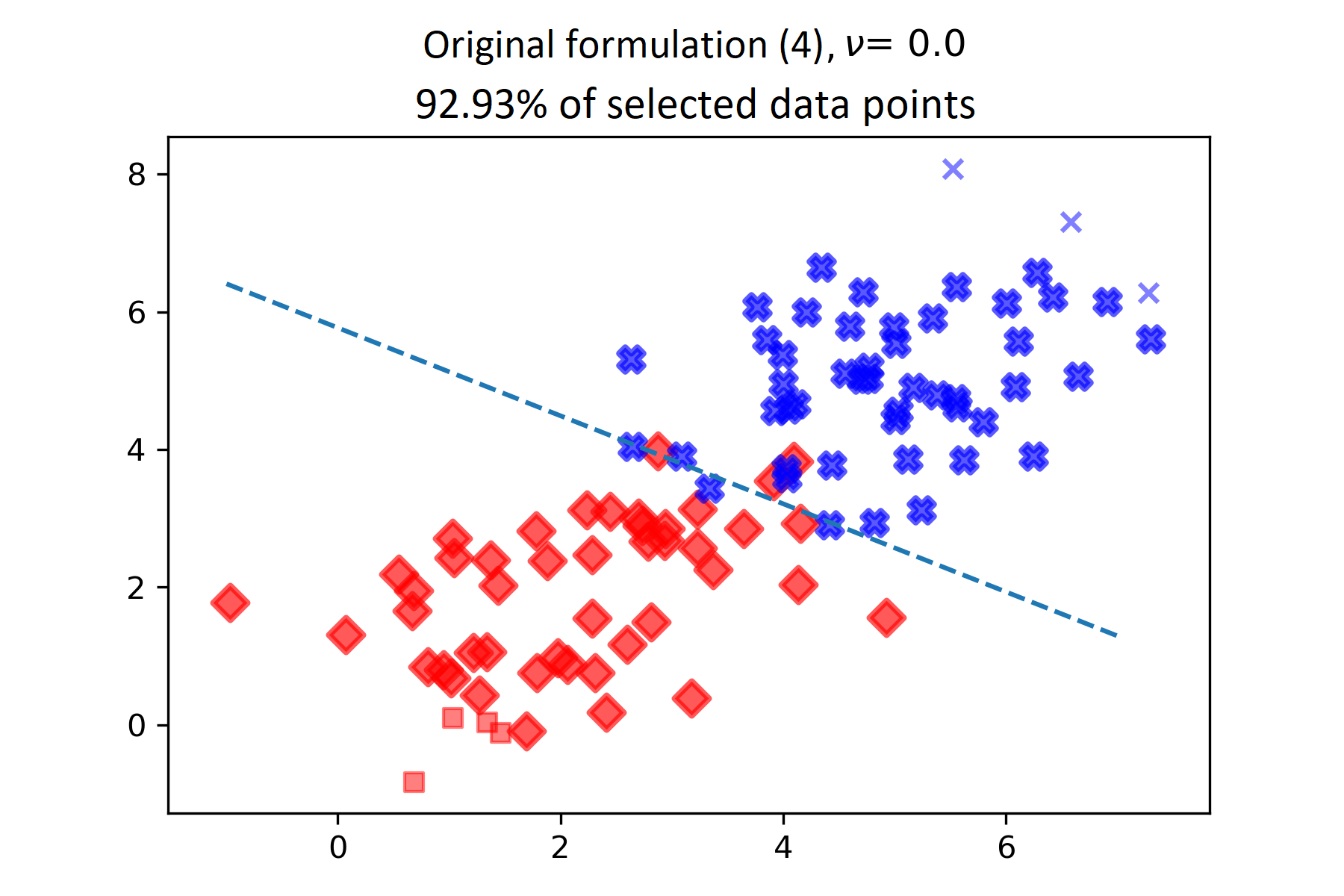}
    \end{minipage}
    \hfill
    \begin{minipage}[b]{0.48\textwidth}
        \centering
        \includegraphics[width=\textwidth]{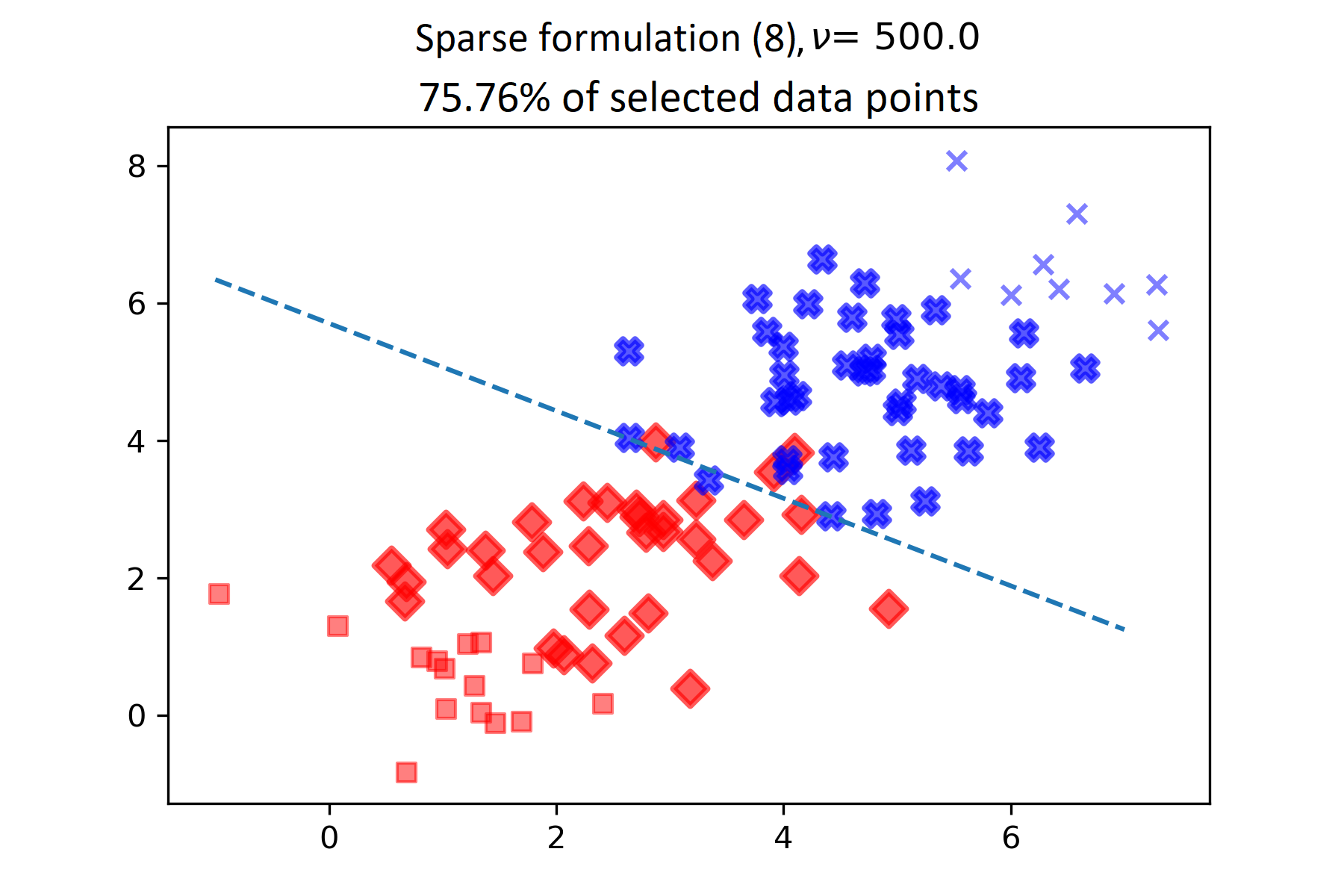}
    \end{minipage}
    \vspace{1em}
    \begin{minipage}[b]{0.48\textwidth}
        \centering
        \includegraphics[width=\textwidth]{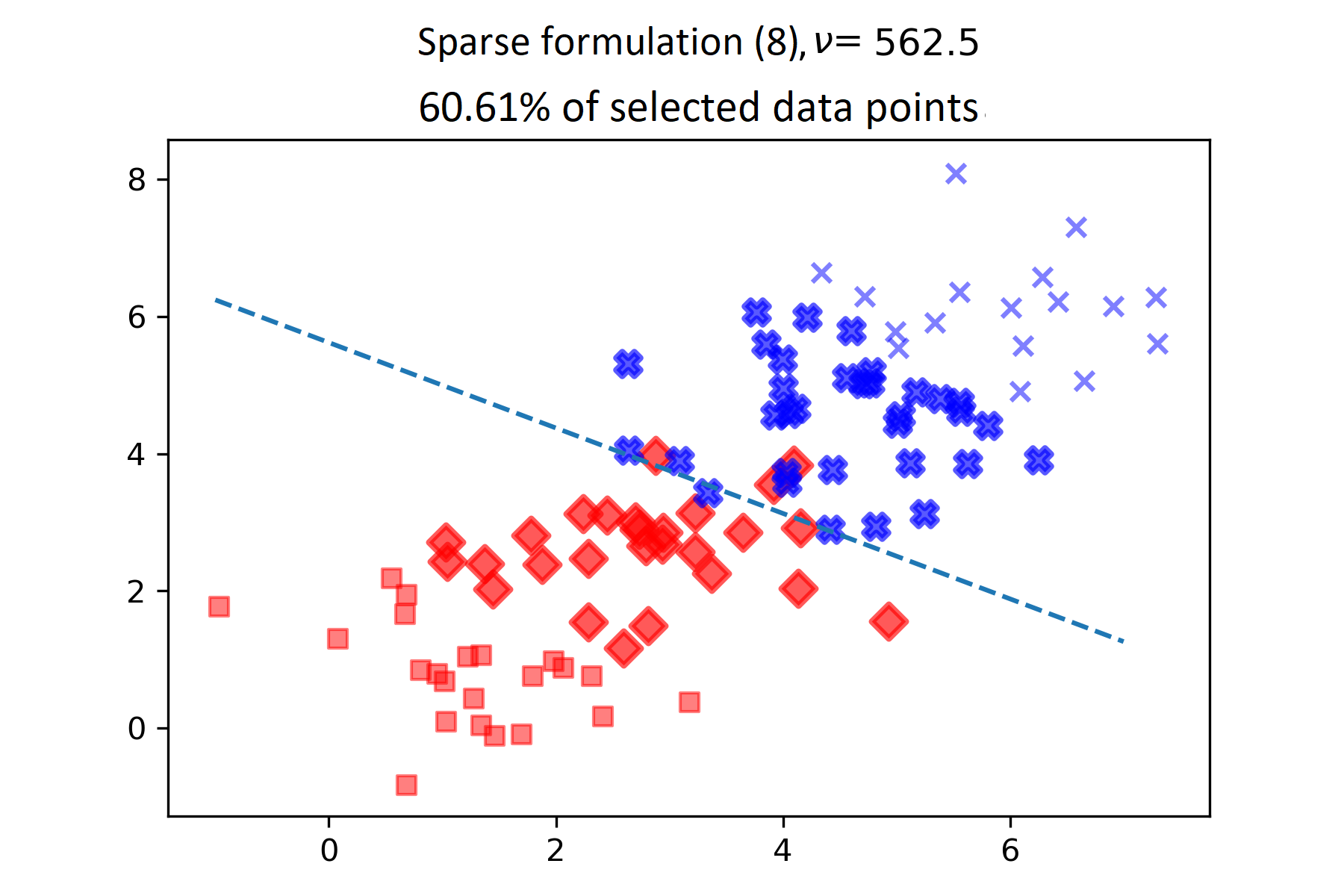}
    \end{minipage}
    \hfill
    \begin{minipage}[b]{0.48\textwidth}
        \centering
        \includegraphics[width=\textwidth]{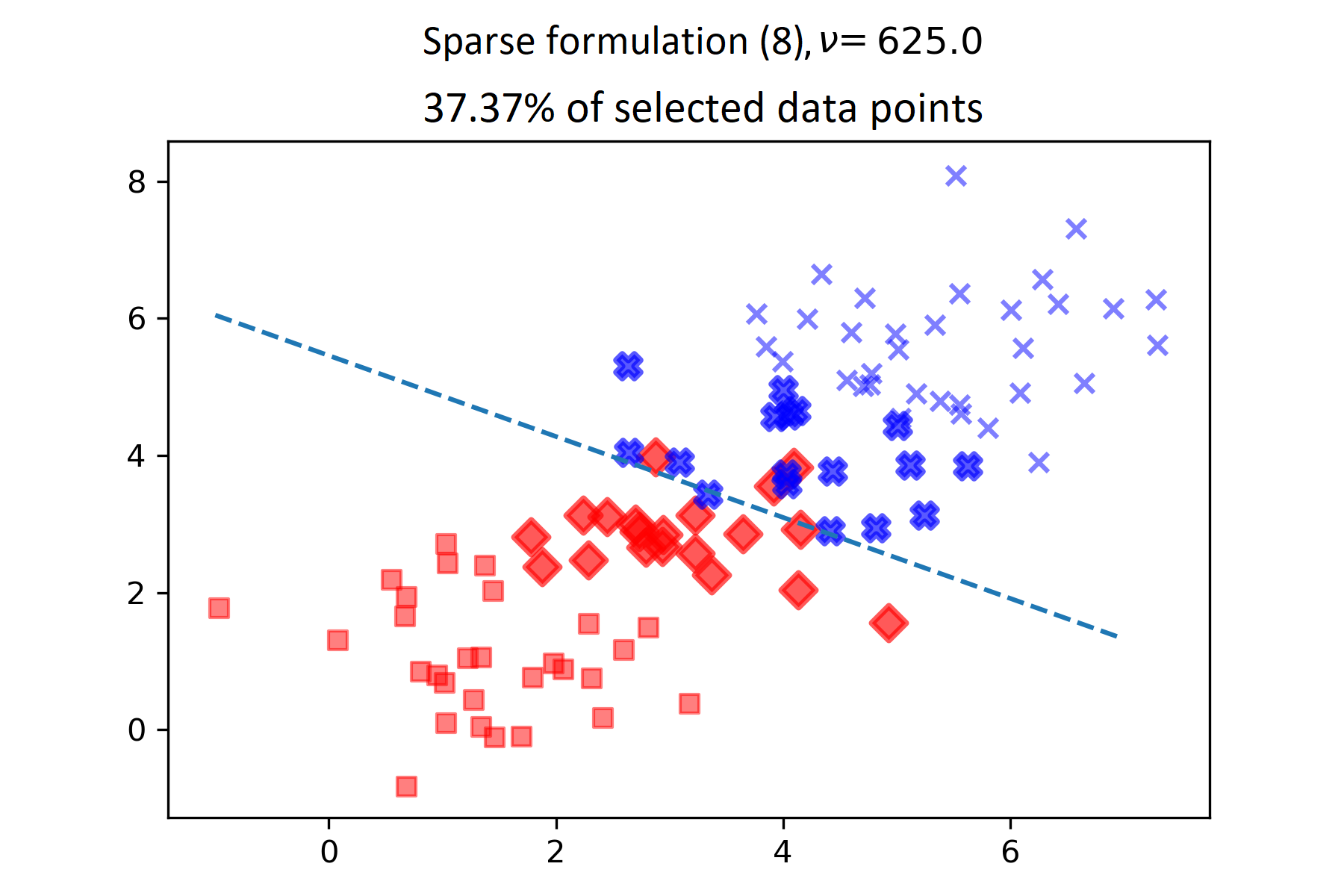}
    \end{minipage}
    \caption{Scatterplots of the \blue{Synth} dataset with 2 classes (red and blue), $p=2$ and $N=99$, for four values of the hyperparameter $\nu \in \{0,500,562.5,625\}$. \blue{Setting $\nu=0$ of the top-left plot amounts to solving the original formulation \eqref{eq:k-primal}.} \textcolor{black}{Data points selected by sparsity-inducing formulation (with nonzero $\alpha_i$) are} represented with bold red rhombuses and bold blue crosses, \textcolor{black}{discarded/excluded} data points are represented with thin blue crosses and red small squares.}\label{fig:explanation1}
\end{figure}

Figure \ref{fig:explanation1} provides a visualization of the impact of the variable $\rho$ in the formulation \eqref{eq:sparse-primal} for the \blue{Synth} dataset with $C=60$ through four scatterplots. \blue{As the value of $\nu$  increases, ranging from $\nu=0$ (which amounts to solve the original formulation \eqref{eq:k-primal}) to $\nu=625$, the number of selected data points decreases from the $92.93\%$ to the $37.37\%$, where the} \textcolor{black}{selected} \textcolor{black}{data points are those associated to $\alpha_i$ values larger than a threshold set to $1e^{-5}$. It is worth pointing out that not only the number of \red{selected} data points substantially decreases but also the focus is progressively restricted to those that are closer to the decision boundary.}

\begin{figure}
    \centering
    \includegraphics[width=0.7\linewidth]{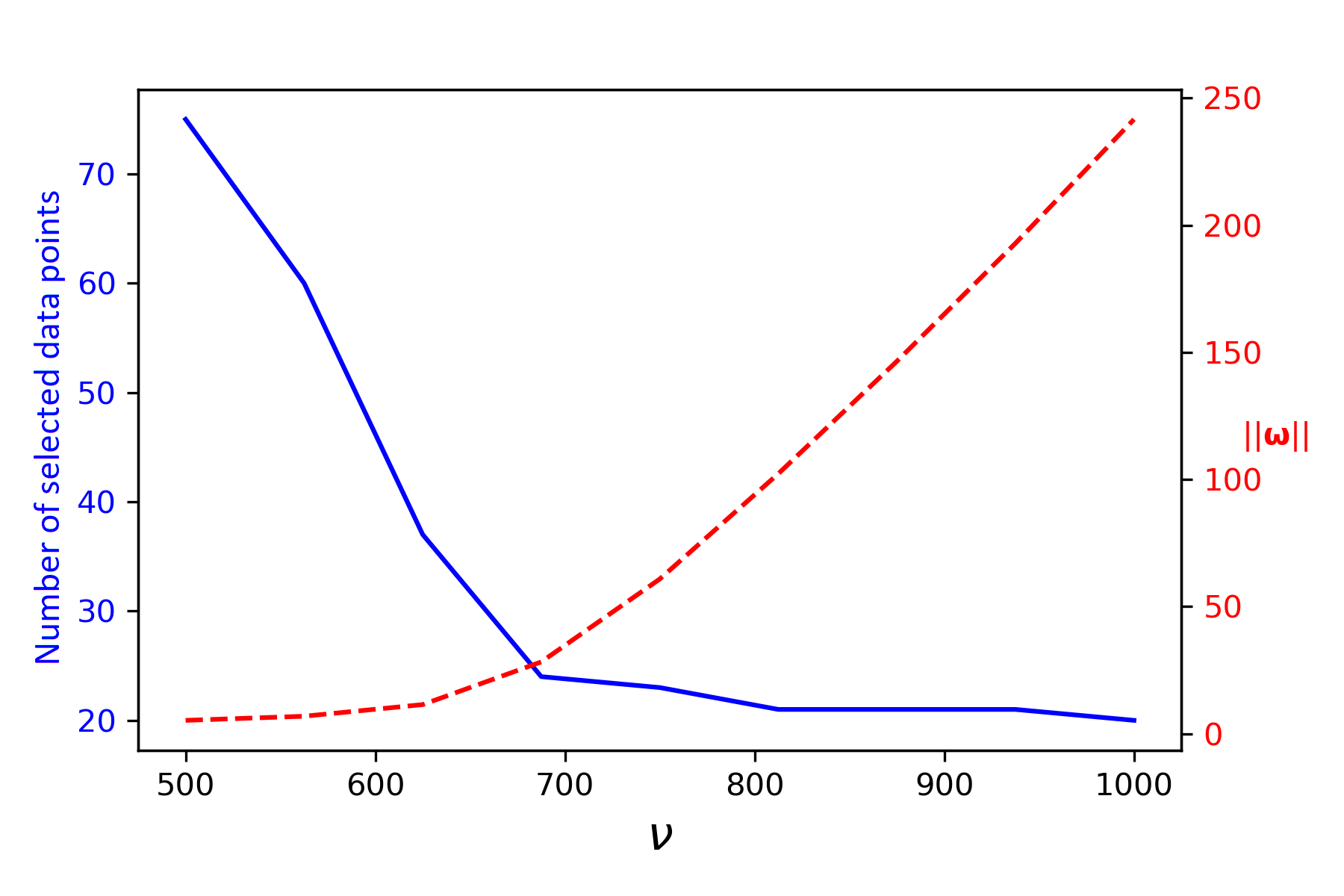}
    \caption{\blue{Plots of the number of data points selected by the sparsity-inducing formulation (in blue) and of the norm of the vector $\bm{\w}$ (in red) as functions of the hyperparameter $\nu$.}}
    \label{fig:explanation2}
\end{figure}

Figure \ref{fig:explanation2} indicates how the number of data points selected by formulation \eqref{eq:sparse-primal} and the norm of the vector $\bm{\w}$ vary when the value of $\nu$ increases from $500$ to $1000$ in 8 equally spaced steps.  The x-axis corresponds to the $\nu$ values, while the left y-axis corresponds to the number of selected data points (blue curve), and the right y-axis to the norm of the decision boundary coefficient vector $\|\bm{\w}\|$ (red curve). 
\blue{Clearly, as the value of $\nu$ grows the presence of the $\rho$ term in the KLR primal formulation \eqref{eq:sparse-primal} induces progressively sparser solutions}, decreasing the number of \textcolor{black}{selected} data points down to about $20\%$. Larger values of $\nu$ also leads to an increase in the norm of $\bm{\w}$ due to the reduced impact of the $\|\bm{\w}\|^2$ term in the objective function.

\subsubsection{Differences with respect to thresholding heuristics}

\blue{As \textcolor{black}{highlighted} above, the increase in $\nu$ leads to higher values of $\alpha_i$  related to the data points closer to the decision boundary and lower values for $\alpha_i$ associated with well-classified points further away from the boundary. Since the introduction of the $\nu \rho$ term in our sparse KLR formulation \eqref{eq:sparse-primal} implies that the larger the value of hyperparameter $\nu$ the larger $\sum\limits_{i \in I}\alpha_i$ (see the KKT conditions \eqref{eq:final_c2}), then the sum of the dual variables $\alpha_i$ corresponding to the data points which lie closer to the decision boundary increases.} 


It is worth pointing that this is not equivalent to applying in the original KLR dual formulation \eqref{eq:k-dual} a thresholding method to sparsify the trained model. \blue{A natural way to define a thresholding heuristic after the training phase is to set to zero all dual variables $\alpha_i$ below a certain threshold. Notice that, by considering the connection between dual and primal variables, namely  $\xi_i= -y_{i}(\bm{\w}\cdot z_{i}-b)$ and $ \xi_i(\alpha_i) = g'^{-1}(\frac{\alpha_i}{C})$ (see \eqref{eq:k-primal-con} and \eqref{eq:pr-du}), this is equivalent to set to zero all $\alpha_i$ associated to data points whose distance from the  decision boundary is larger than a certain value.}


We implemented two versions of the thresholding heuristic to illustrate the difference with respect to our approach. In the first version, we solve the (non-sparse) primal formulation \eqref{pr:pr1} (with $\nu=0$) and construct the decision boundary by considering only the $\alpha_i$ exceeding a threshold that guarantees the same level of sparsity obtained by solving our (sparse) formulation \eqref{eq:sparse-primal} for a given $\nu > \bar{\nu}$. In the second version, we remove from the dataset those data points whose corresponding $\alpha_i$ have been set to zero according to the first version of the heuristic, and we train again the KLR model solving formulation \eqref{pr:pr1} with respect to the reduced dataset. \textcolor{black}{Note that the second version of the heuristic does not exploit at all information related to easy-to-classify data points, since it only considers the data points selected by the first version of the heuristic. For our illustrative example, the second version of the heuristic yields a naive classifier that assigns the same class label to all data points. Hence, we briefly discuss the second version below, without reporting the results.}

For the Synth dataset we solve formulation \eqref{eq:k-primal} with $C=60$, by setting $\nu=0$ for the non-sparse solution and $\nu=1100$ for the sparse one. For $\nu=1100$, the optimal solution selects 20 data points, while, for $\nu=0$, 92 data points are considered. To guarantee the same level of sparsity we select from the optimal solution obtained with $\nu=0$ the 20 data points with higher $\alpha_i$s.

\begin{figure}
    \centering
    \includegraphics[width=0.8\textwidth]{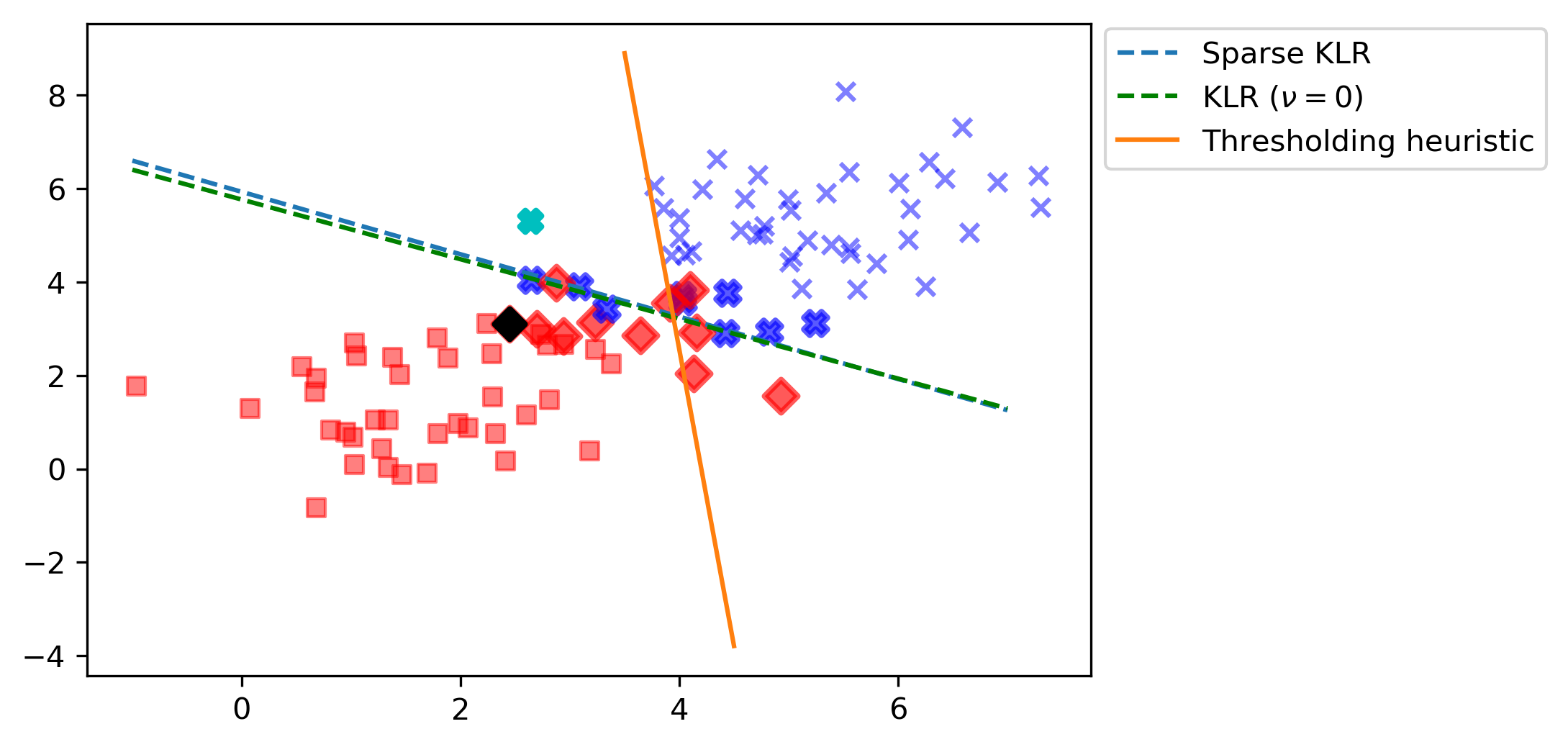}
    \caption{Scatter plot of the \blue{Synth} dataset. The three lines represent the decision boundary related to Sparse KLR (blue dashed line), \textcolor{black}{Thresholding} heuristic (orange solid line), and the KLR formulation with $\nu=0$ (green dashed line). The rhombuses and crosses represent the \textcolor{black}{selected} data points for the red and blue classes, respectively. The cyan data point is selected solely by the sparse formulation while the black one is selected by imposing the threshold.}
    \label{fig:first_s}
\end{figure}

\textcolor{black}{Figure \ref{fig:first_s} shows a scatter plot of the Synth dataset, where class membership is represented by color (red and blue). The linear classifiers produced by Sparse KLR (depicted as a blue dashed line) and the thresholding heuristic (shown as an orange solid line) are shown.} Data points represented with bold rhombuses and crosses are those \textcolor{black}{selected} for classification by both our Sparse KLR and the thresholding heuristic. 
\textcolor{black}{It is interesting to note that the two methods select different data points. In particular, 
the cyan data point is only selected  by the Sparse KLR, while the black data point is only selected by the thresholding heuristic.} Moreover, not only the two hyperplane classifiers differ substantially but also the one obtained with the sparse formulation turns out to be more accurate ($95\%$ accuracy vs. $92\%$). Indeed, the Sparse KLR classifier tends to be very similar to the one obtained by training the non sparse KLR formulation (depicted in Figure \ref{fig:first_s} as a green-colored dashed line) with the same $C=60$, which is the best possible in terms of training accuracy.

\begin{figure}
    \centering
\includegraphics[width=0.8\textwidth]{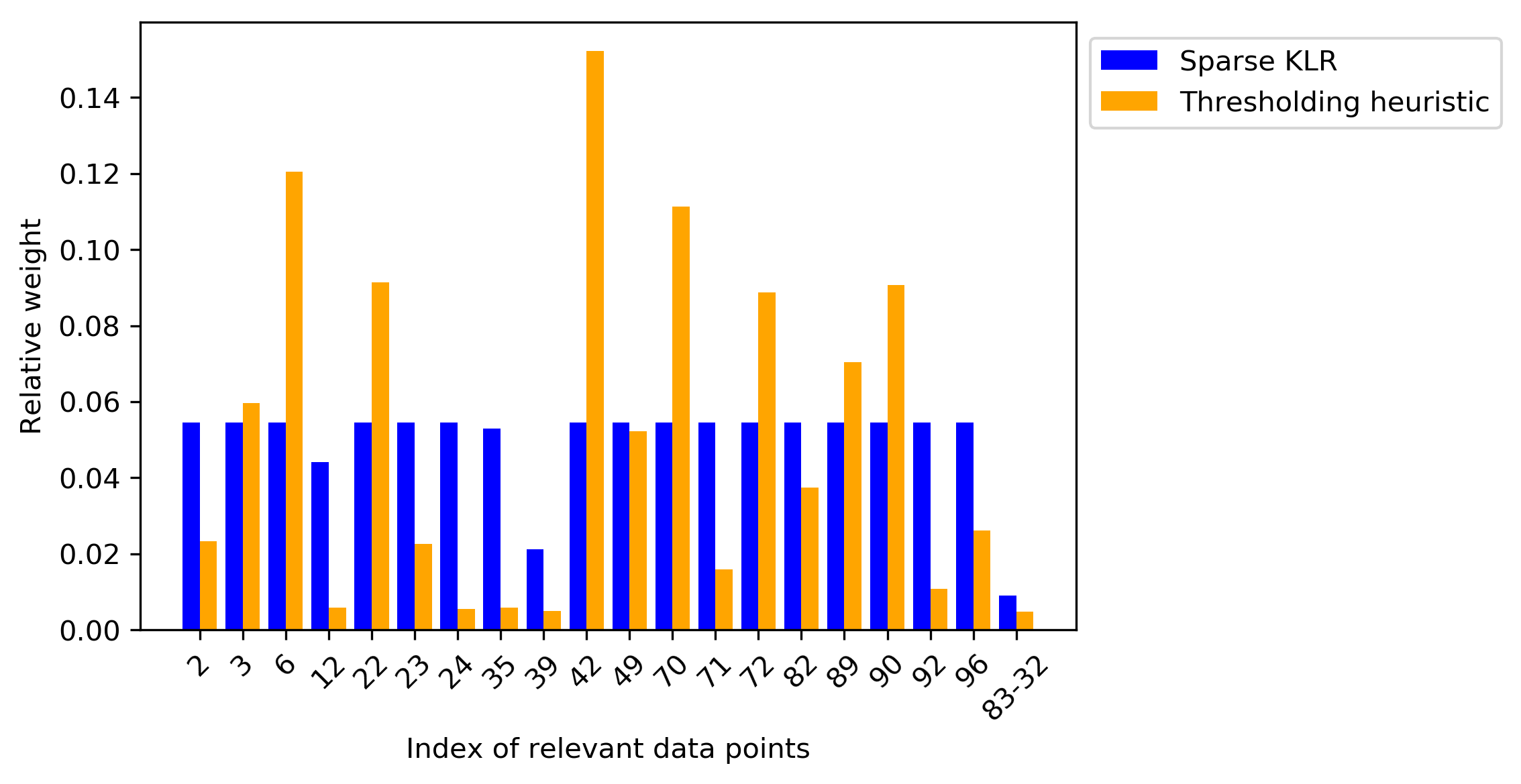}
    \caption{Histograms representing the relative weights ($\alpha_i / \sum_{i \in I} \alpha_i$) of \textcolor{black}{selected} data points in the solutions provided by the Sparse KLR formulation (blue) and the thresholding heuristic (yellow).}
    \label{fig:hist}
\end{figure}

Another interesting aspect influencing the difference between the two classifiers is the relative weight of each $\alpha_i$ with respect to the sum of all other selected data points, namely $\alpha_i / \sum_{i \in I} \alpha_i$. Figure \ref{fig:hist} shows histograms with the relative weights of the 20 data points (whose indices are reported in the x-axis) selected by the sparse formulation (in blue) and the thresholding heuristic (in yellow). \textcolor{black}{In our sparse formulation, the relative weights of the selected data points are more evenly distributed 
\textcolor{black}{than} those obtained by the thresholding heuristic. }
\blue{In summary, the sparse \textcolor{black}{KLR} formulation and the thresholding heuristic yield two distinct decision boundaries because they select different \textcolor{black}{subsets} of data points and lead to different distributions of the relative weight associated to the corresponding dual variables.}

A second possible version of the thresholding heuristic consists in training the non sparse KLR formulation on the restricted dataset containing only the data points identified by the first thresholding heuristic. When comparing our sparse KLR classifier with that of the second thresholding heuristic, we notice that the latter achieves a substantially lower accuracy. 
This is because, when solving formulation \eqref{eq:k-primal} considering only the 20 selected data points, we focus only on the data points that are either misclassified or very close to the boundary, without taking into account the easily classifiable data points. We do not report this classifier in the figure since it classifies all the data points in the same class.

\section{A decomposition method \textcolor{mygreen}{for training} sparse binary KLR}\label{sec:dec}

Decomposition 
\textcolor{mygreen}{methods} are widely used in training 
\textcolor{mygreen}{ML models} whenever, due to dimensionality, standard optimization methods involving first or \textcolor{mygreen}{second order} information are out of reach (see e.g., \cite{platt1998sequential,lucidi2009convergent,manno2018parallel,manno2016convergent}). 
\textcolor{mygreen}{They} can be designed to leverage the underlying structure of the training problems, with positive impacts on the computational efficiency and on the ability to escape from \textcolor{mygreen}{poor} quality solutions (see e.g., \cite{grippo2015decomposition,amaldi2023multivariate,consolo2025}).

Decomposition algorithms iteratively split the original problem into a sequence of smaller subproblems in which, \textcolor{black}{at each iteration,} only a subset of variables, the so-called working set, is involved in the optimization process, while the other ones are kept fixed to their current values. The choice of the variables to be included, commonly denoted as working set selection, is crucial for the computational efficiency and for the theoretical convergence properties.

In the next subsection, we present a SMO-type decomposition method for 
formulation \eqref{eq:2BM}, and we state its asymptotical global convergence property which is proved in \ref{subsec:prop_app}.

\subsection{A SMO-type 
decomposition algorithm \textcolor{mygreen}{based on second-order information}}\label{subsec:SMO}
The structure of the dual KLR formulation \eqref{eq:2BM} is amenable to \textcolor{mygreen}{decomposition}. 
In particular, as similarly noticed in \cite{keerthi2005fast} for formulation \eqref{eq:k-dual}, \eqref{eq:2BM} allows for the adaptation to KLR of the efficient SMO-type decomposition algorithms 
\textcolor{mygreen}{developed} for SVMs.

In SMO-type algorithms, at each iteration $k$ the working set involves only two variables, whose indices are referred to as $\{i^*,j^*\} \subseteq \{1,\ldots,N\}$ \footnote{For simplicity of notation, the dependence of $\{i^*,j^*\}$ from iteration $k$ is omitted.}.  Notice that, given a current feasible solution $\bm\alpha$ and the working set indices $\{i^*,j^*\}$, due to equality constraint \eqref{eq:final_c}, the only feasible directions for variables $\alpha_{i^*}$ and $\alpha_{j^*}$ are, respectively, of the form $\frac{t}{y_{i^*}}$ and $ -\frac{t}{y_{j^*}}$, where $t$ represents the steplength. Therefore, the two-dimensional SMO subproblem can be recast into a one-dimensional problem with respect to the steplength variable $t$ as follows
\begin{subequations}\label{eq:onedim_pr}
	\begin{equation}\label{eq:subprob_obj}
		\min_{t} \quad \phi_{i^*,j^*}(t) = \frac{1}{2} \| \bm{\w}( \tilde{\bm{\alpha}}(t) )\|^2 +C\sum_{i\in I}G(\frac{\tilde{\alpha}_i(t)}{C})- \lambda \sum_{i\in I} \tilde{\alpha}_i(t)
	\end{equation}
	\begin{equation}\label{eq:subprob_c1}
		\text{s.t.} \qquad   \overline{0} \leq \tilde{\alpha}_{i^*}(t) \leq \overline{C}
	\end{equation}
	\begin{equation}\label{eq:subprob_c2}
		\qquad \quad \,\, \overline{0} \leq \tilde{\alpha}_{j^*}(t) \leq \overline{C},
	\end{equation}
\end{subequations}
where  $ \tilde{\alpha}_{i^*}(t) = \alpha_{i^*} + \frac{t}{y_{i^*}}, \, \tilde{\alpha}_{j^*}(t) = \alpha_{j^*} - \frac{t}{y_{j^*}} $ and $\tilde{\alpha}_{s}(t) = \alpha_{s}$ 
\textcolor{black}{for all $s$ with $s \neq i^*$ and $s \neq j^*$}.

Before describing the proposed working set selection \textcolor{mygreen}{procedure}, we introduce the concept of ``violating pair", consisting of a pair of indices $\{i,j\}$ whose associated variables $\alpha_i$ \textcolor{mygreen}{and} $\alpha_j$ violate together \textcolor{black}{the} optimality conditions \eqref{eq:opt_cond}. A violating pair, which is a good candidate to be included in the working set, can be determined by selecting one index $i \in I_{up}$ and one index $j \in I_{low}$ such that 
\begin{equation}\label{eq:violating}
	-y_i \nabla f(\bm{\alpha})_i > -y_{j} \nabla f(\bm{\alpha})_{j}.
\end{equation}

Among all possible violating pairs, consider $\{i^{MVP},j^{MVP}\}$ associated to the maximal violation of the optimality conditions \eqref{eq:opt_cond}, the so called ``Maximal Violating Pair" (MVP). The MVP can be easily determined (at the cost of ordering elements $-y_i \nabla f(\bm{\alpha}_i)$) as

\begin{equation}
\label{eq:MVP_def}
  i^{MVP} = \text{arg} \max_{i \in I_{\text{up}}(\bm{\alpha})} -y_i \nabla f(\bm{\alpha}_i) \; \; \text{and}\; \;  j^{MVP} = \text{arg} \min_{j \in I_{\text{low}}(\bm{\alpha})} -y_j \nabla f(\bm{\alpha}_j).   
\end{equation}

\textcolor{mygreen}{In the SMO-type decomposition method for KLR proposed in \cite{keerthi2005fast}, the working set selection procedure consists in iteratively selecting the MVP according to first-order information.}


Inspired by the computationally more efficient working set selection \textcolor{mygreen}{procedure} 
proposed in \cite{fan2005working} for SVMs, we devise for the KLR dual formulation \eqref{eq:2BM} a working set selection procedure which 
takes advantage of \textcolor{mygreen}{second-order} information. \textcolor{black}{After setting the first index $i^*$ as $i^{MVP}$, one selects,} among all possible candidate indices $j \in I_{\text{low}}(\bm{\alpha})$ satisfying \eqref{eq:violating}, 
\textcolor{black}{the index}
$j^*$ 
for which the quadratic approximation of the objective function \eqref{eq:subprob_obj} of the sub-problem 
\textcolor{black}{associated to}
$\{i^*,j^*\}$ is minimized.
In particular, for a given pair of indices $\{i^*,j\}$ we approximate 
\textcolor{black}{the}
objective function \eqref{eq:subprob_obj} of the corresponding sub-problem, with its \textcolor{mygreen}{second-order} Taylor expansion in a neighborhood of $t=0$, as
\begin{equation}
	\label{eq:approx}
	\phi_{i^*,j}(0) + \phi_{i^*,j}^{'}(0)t+ \frac{1}{2}\phi_{i^*,j}(0)^{''}t^2. \end{equation}
Then, \textcolor{black}{the index} $j^{*}$ is selected as the one that minimizes \textcolor{black}{\eqref{eq:approx}} the most. The minimizer of \eqref{eq:approx} can be obtained analytically. Indeed, for a given index $j$, the $t$ value minimizing \eqref{eq:approx}, 
\textcolor{mygreen}{denoted by} $t^*$, is $t^*= -\frac{v_{i^*j}}{2q_{i^*j}}$ and the corresponding objective function value is $-\frac{v_{i^*j}^2}{2q_{i^*j}}$ (up to 
\textcolor{black}{a}
constant term), where 
\begin{equation}\label{eq:b}
	\begin{split} v_{i^*j}=  \sum_{s\in I} \alpha_s y_s Ker(\bm{x}_j,\bm{x}_s) + y_jG^{'}(\frac{\alpha_j}{C}) -\lambda y_j - \\ \sum_{s\in I} \alpha_s y_s Ker(\bm{x}_{i^*},\bm{x}_s) - y_{i^*}G^{'}(\frac{\alpha_{i^*}}{C}) +\lambda y_{i^*}
	\end{split}
\end{equation}
and 
\begin{equation}\label{eq:a}
	\begin{split}q_{i^*j}=Ker(\bm{x}_{i^*},\bm{x}_{i^*})+Ker(\bm{x}_{j},\bm{x}_{j}) - 2Ker(\bm{x}_{i^*},\bm{x}_{j}) +\\ \frac{C}{\alpha_{i^*}(C-\alpha_{i^*})} + \frac{C}{\alpha_{j}(C-\alpha_{j})}.
 \end{split}
\end{equation}

\noindent 
It is important to note that \textcolor{mygreen}{the solution} $t^*$ minimizing problem \eqref{eq:approx} is used exclusively for selecting the second index $j^*$ in the \textcolor{mygreen}{working set selection procedure}. Once the pair $(i^*,j^*)$ is determined, the update of the dual variables $(\alpha_{i^*},\alpha_{j^*})$ is performed by solving subproblem \eqref{eq:onedim_pr}.

To summarize, we propose the following \textcolor{mygreen}{second-order} working set selection \textcolor{mygreen}{procedure}:
\bigskip

\noindent \textbf{Second-order Working Set \textcolor{black}{Selection} 
	 (WSS)}
\medskip
\hrule
\begin{itemize}
    \item[1)] \textcolor{black}{Select $i^* = \arg\max\limits_i \{ -y_i \nabla f(\bm{\alpha})_i\,|\, i \in I_{\text{up}}(\bm{\alpha}) \} = \arg\max\limits_i \{ -\sum\limits_{j\in I} \alpha_j y_j Ker(\bm{x}_i,\bm{x}_j) - y_iG^{'}(\frac{\alpha_i}{C}) - \lambda y_i\,|\, i \in I_{\text{up}}$\}}
	\item[2)] Select $ j^* = \arg\min\limits_j \{-\frac{v_{i^*j}^2}{q_{i^*j}}\,|\, j \in I_{\text{low}}(\bm{\alpha}), -y_j \nabla f(\bm{\alpha})_j < -y_{i^*} \nabla f(\bm{\alpha})_{i^*}   \} $,
	where $v_{i^*j}, q_{i^*j}$ are as in \eqref{eq:b} and \eqref{eq:a}

	\item[3)] Return the pair $\{i^*,j^*\}$
\end{itemize}
\hrule
\bigskip

\noindent 
\textcolor{black}{and the SMO-type decomposition training algorithm described in SMO Algorithm}.
\bigskip

\begin{algorithm}[H]
\caption{Sequential Minimal Optimization decomposition algorithm}\label{alg:smo}
\begin{algorithmic}[1]
        \State Choose $\bm{\alpha}^0$ feasible starting solution and set $k = 0$
        \If {$\bm{\alpha}^{k}$ satisfies \textcolor{black}{\eqref{eq:opt_cond}}}
        \State $\bm{\alpha}^* = \bm{\alpha}^{k}$
        \Else 
        \State Select $\{i,j\}$ according to WSS procedure;
        \State Determine $\bm{\alpha}^{new}$ as the optimal solution of \textcolor{black}{\eqref{eq:onedim_pr}};
        \State Let $\bm{\alpha}^{k+1} = \bm{\alpha}^{new}$, set $k = k+1$ and go back to 2
        \EndIf
        
        \State \textbf{return} $\bm{\alpha}^*$
\end{algorithmic}
\end{algorithm}

\vspace{15pt}
For any infinite sequence generated by \hyperref[alg:smo]{SMO Algorithm} we can establish global asymptotic convergence. The convergence analysis, which is reported in \ref{subsec:prop_app}, \textcolor{mygreen}{is an adaptation of the one for the second-order SMO-type method presented in \cite{chen2006study} for SVM. Since due to the presence of the term $C\sum\limits_{i\in I}G(\frac{\alpha_i}{C})$ the objective function \eqref{eq:final} is convex but not quadratic (like for SVM), we consider a quadratic approximation in the WSS.}

The convergence result is as follows.

\begin{proposition}
	\label{pr:pr1}
	Assume \textcolor{black}{that} the kernel matrix $K$ is positive semidefinite. Let $\{\alpha^{k}\}$ be the infinite sequence generated by the SMO-type method Algorithm \ref{alg:smo}. Then $\{\alpha^k\}$ globally converges to the unique optimal solution of problem \eqref{eq:2BM}. 
\end{proposition}

It is worth pointing out that the \textcolor{mygreen}{assumption} 
in Proposition \ref{pr:pr1} of positive \textcolor{mygreen}{semidefiniteness of} kernel matrix $K$ (i.e., satisfying the Mercer’s Theorem \cite{mercer1909xvi}) is very mild from a practical point of view. Indeed, several kernel functions usually adopted in practical applications (e.g., linear, polynomial, \textcolor{mygreen}{Gaussian, and Laplacian} kernels) induce positive semidefinite (or definite) kernel matrices.

\section{\textcolor{black}{Computational experiments}}\label{sec:experiments}

\textcolor{black}{In this section, we first compare the Sparse KLR training formulation \eqref{eq:2BM} solved to optimality via \hyperref[alg:smo]{SMO Algorithm}  (referred to as S-KLR), with IVM, $\ell_{1/2}$-$\text{KLR}$, and SVM on a collection of 12 benchmark datasets in terms of testing accuracy and sparsity. } \textcolor{black}{Then, a second set of experiments is carried out to assess independently the impact of the proposed formulation variant in terms of sparsity, and of the second order WSS procedure in terms of CPU time reduction.}




\subsection{\textcolor{black}{Experimental settings and datasets}}\label{subsec:settings}
S-KLR is implemented in {\tt Python 3.9.7}, using the {\tt Cython} extension to convert the \hyperref[alg:smo]{SMO Algorithm} part of the code in C in order to improve \textcolor{black}{efficiency. For SVM training}, the \texttt{scikit-learn} 1.1.2 package is used. To train IVM, the Import Vector Machine Classifier software \cite{roscher2012i2vm,roscher2012incremental} (version 4.4) in {\tt Matlab 2022b} is adopted. As to the $\ell_{1/2}$-$\text{KLR}$, the half thresholding has been implemented in {\tt Python 3.9.7} using the \texttt{jax 0.4.23} package for the efficient calculation of the gradient. \textcolor{black}{The experiments are carried out on a PC with Intel(R) Core(TM) i7-11370H CPU @ 3.30GHz with 16 GB of RAM.}

To initialize S-KLR we follow the approach outlined in \cite{keerthi2005fast}. For \hyperref[alg:smo]{SMO Algorithm}, it is essential to guarantee that the starting \textcolor{black}{solution} $\bm\alpha^0$ is feasible with respect to constraints \eqref{eq:final_c} and \eqref{eq:final_compact}. By denoting with $m_1$ and $m_2$ the number of \textcolor{black}{data points} in the first and second class, respectively, if $\max\{m1,m2\}<\tau^{-1}$, then 
$\bm{\alpha}^0$ components $\alpha_i^0$ can be set to $\frac{1}{m_1}$ if the $i$-th data point is in the first class and $\frac{1}{m_2}$ otherwise. In the unlikely case in which $\max \{m1,m2\}>\tau^{-1}$, then if $m_1>m_2$ 
we can set $\alpha_i=\overline{0}$ for $i$ in the first class, and $\alpha_i = \frac{m_1\overline{C}}{m_2}$ otherwise, provided \textcolor{black}{that}
$\frac{m_1+m_2}{m_2}\tau\leq C$. 

Two alternative stopping criteria are considered for \hyperref[alg:smo]{SMO Algorithm}: a maximum number of iterations (\textcolor{black}{set} to 10000), and the satisfaction of \textcolor{black}{the} optimality conditions \eqref{eq:opt_cond} up to a certain tolerance $\tau = 1e^{-5}$. As for the parameter $\gamma$ \textcolor{black}{appearing in the definition of the bounds $\overline{0}$ and $\overline{C}$,
we set} $\gamma = 1e^{-5}$. 

\textcolor{black}{Notice that, due to the auxiliary bounds of constraints \eqref{eq:final_compact}, no optimal dual variables will assume exactly the value zero, as every component $\alpha_i$ of any feasible solution must satisfy $\alpha_i\geq \bar0= \gamma>0$. However, in the reported experiments the components $\alpha_i^*$ of the optimal solution $\bm{\alpha}^*$ whose values reach the lower bound ($\alpha_i^*=\gamma$)  are considered equal to zero and are not involved in the model for prediction.}

Subproblem \eqref{eq:onedim_pr} is solved with the ``Newton-CG" algorithm (see e.g., \cite{nocedal1999numerical}) implemented in the {\tt Scipy 1.9.0 Python} package. 
\begin{table}[h]
	\centering
	\caption{Description of 12 datasets}\label{tab:data}
	\begin{tabular}{llcc}
		\hline
		Dataset                 & \multicolumn{1}{c}{Abbreviation} & N                       & p                      \\ \hline
		banknote                & banknote                         & 1372                    & 4                      \\
		coil2000                & coil2000                         & 9822                    & 85                     \\
		diabetes                & diabetes                         & 768                     & 8                      \\
		ionosphere              & ionosphere                       & 351                     & 34                     \\
		magic                   & magic                            & 19020                   & 10                     \\
		monk2                   & monk2                            & 432                     & 6                      \\
		ringnorm                & ring                             & 7400                    & 20                     \\
		sonar                   & sonar                            & 208                     & 60                     \\
		spambase                & spambase                         & 4597                    & 57                     \\
		twonorm                 & twonorm                          & 7400                    & 20                     \\
		waveform                 & waveform                         & 5000                    & 21                     \\
		wisconsin-breast-cancer & wisconsin                        & 569 & \multicolumn{1}{l}{30} \\ \hline
	\end{tabular}
\end{table}
\textcolor{black}{Numerical} experiments are conducted on 12 datasets from the UCI Machine Learning Repository \cite{asuncion2007uci} and the LIBSVM repository \cite{chang2011libsvm} \footnote{\textcolor{black}{The three-class dataset waveform, considered in both \cite{keerthi2005fast} and \cite{zhu2005kernel}, has been transformed into a binary classification problem where the data points in class 1 must be separated from those in classes 0 and 2. It is worth noting that the separation of the data points in} class 0 or 2 does not significantly affect the overall results.}. The number of data points $N$ in each dataset ranges from $208$ to $19020$, while the number of features $p$ ranges from $4$ to $85$ (see Table \ref{tab:data} for details). For all datasets, all features are scaled in the interval $[0, 1]$, \textcolor{black}{and} the Gaussian kernel $K(\bm{x}_i,\bm{x}_j)=\text{exp}(-\frac{\| \bm{x}_i - \bm{x}_j\|^2}{2\sigma^2}) $ with $\sigma = 1$ is adopted. 

The testing accuracy and sparsity \textcolor{black}{of the ML models obtained using} all the compared methods are evaluated through k-fold cross-validation, with $k = 5$. 
Since the values of hyperparameters $C$ and $\lambda$ can significantly affect the model performance, we conducted a grid search to determine good estimates. \textcolor{black}{For every fold and model, the values of $C$ and $\lambda$ were selected using a validation set consisting of $5\%$ of the training data points. After the selection of both hyperparameters, the model was retrained on the entire training set with the chosen values.} 
For all methods, both performance measures are computed by varying the regularization parameter $C$ in the interval $\{10^r: -4 \leq r \leq 4, r \in \mathbb{Z}\}$. For S-KLR, a further hyperparameter $\lambda$ related to the sparsity inducing term must be set. For a given regularization value $C$, $\lambda$ varies 
\textcolor{black}{over}
10 equally spaced values in the interval $\left[0,C\right]$ (including $0$ and $C$).

Although the above-mentioned grid search for S-KLR over both $C$ and $\lambda$ is computationally heavier than in the presence of a single hyperparameter (like for SVMs), in \ref{subsec:c-lambda} we provide experimental evidence that the $\lambda$ fine-tuning can be avoided. Indeed, for a fixed value of $C$, 
$\lambda$ can be set to a relatively small value depending on $C$, 
without compromising S-KLR performance. 

Concerning $\ell_{1/2}$-$\text{KLR}$, in \cite{xu2013sparse} the authors adapt an approach proposed originally in \cite{xu2012l_} for which the value of the hyperparameter $C$ is automatically \textcolor{black}{selected} based on a predefined level of sparsity. They assume that a priori information on a ``proper" sparsity level is available and do not carry out a grid search on the $C$ value. In this work, we consider the general setting where no prior information is available and we apply in the $\ell_{1/2}$-$\text{KLR}$ experiments the same standard grid search used for S-KLR, IVM and SVM.

\subsection{\textcolor{black}{Numerical results}}\label{subsec:results}
\textcolor{black}{We have carried out two types of experiments. First, we compare the performance of the proposed S-KLR formulation and decomposition algorithm with that of IVM, $\ell_{1/2}$-$\text{KLR}$, and SVM in terms of testing accuracy and sparsity.} \textcolor{black}{Second, we investigate how the proposed formulation variant \eqref{eq:2BM} improves model sparsity with respect to the formulation in \cite{keerthi2005fast}, and we assess the impact of the second order WSS procedure in terms of CPU time with respect to a benchmarking version of S-KLR implementing the same first order WSS procedure in \cite{keerthi2005fast}.} 




\textcolor{black}{As to the first type of experiments, Table \ref{tab:results} reports the average testing accuracy and sparsity level across the $k=5$ folds. These results were obtained by selecting the best values of the hyperparameters ($C$ for all methods and also $\lambda$ for S-KLR) through validation over the aforementioned grid of values.}
\textcolor{black}{For S-KLR, $\ell_{1/2}$-$\text{KLR}$, and SVM, the table also reports the average results obtained by selecting the hyperparameters, via validation, that yield the sparsest model among the three most accurate ones. These model versions are referred to as ``S-KLR sparsest of 3 most accurate", ``$\ell_{1/2}$-$\text{KLR}$ sparsest of 3 most accurate", and ``SVM sparsest of 3 most accurate".} It is worth noting that in some cases, such as for S-KLR on \textcolor{black}{twonorm} dataset or for SVM on the spambase dataset, the highest level of sparsity does not necessarily coincide with the best testing accuracy.

The results in Table \ref{tab:results} indicate that both S-KLR and SVM achieve equivalent and substantially higher accuracy levels than IVM and $\ell_{1/2}$-$\text{KLR}$. In particular, S-KLR and SVM yield an average testing accuracy of approximately \textcolor{black}{$92.5\%$, while IVM of $84.9\%$ and $\ell_{1/2}$-$\text{KLR}$ of $86.7\%$.} S-KLR 
attains a good average degree of sparsity \textcolor{black}{($42.18\%$ of data points are used), \textcolor{black}{but SVM yields sparser models} (only $28.8\%$} of data points are support vectors). 
\textcolor{black}{It is worth noting that, when considering the experiments for the sparsest of the $3$ most accurate models, the average testing accuracy of S-KLR still amounts to $92.5\%$, whereas that of SVM decreases to $90.5\%$, and the sparsity gap between S-KLR and SVM narrows to $6.3\%$ ($33.1\%$ and $26.8\%$, respectively). Moreover, when comparing the sparsest model among the $3$ most accurate S-KLR model with SVM, the average sparsity difference further decreases to less than $4.5\%$, while maintaining the same level of accuracy.}


IVM, which is based on a greedy heuristic, tends to achieve remarkable sparsity levels  \textcolor{black}{($5.6\%$)} but at the cost of significantly compromising the testing accuracy (on average $84.9\%$). 
\textcolor{black}{For the banknote, diabetes, ionosphere, magic, monk2, and ring datasets, IVM yields testing accuracies which are lower by at least $6.7\%$ and by up to $21\%$ than those provided by S-KLR and SVM. Note that the standard deviation of the average testing accuracy across all datasets is substantially higher for IVM ($0.096$) than for S-KLR and SVM (about $0.066$).}

$\ell_{1/2}$-$\text{KLR}$ achieves considerably lower testing accuracy compared with S-KLR and SVM (on average \textcolor{black}{$5.8\%$ lower), and considerably lower sparsity level ($67.11\%$} of the data points are used) with respect to S-KLR, 
SVM 
and to IVM. \textcolor{black}{As far as the 3 most accurate model versions are concerned, $\ell_{1/2}$-$\text{KLR}$ yields testing accuracies that are comparable to those provided by S-KLR and SVM for the coil2000 and waveform datasets, whereas it achieves testing accuracies which are at least $6.7\%$ lower for all the other datasets.}

To summarize, S-KLR and SVM achieve a better trade-off between testing accuracy and sparsity with respect to IVM \textcolor{black}{and $\ell_{1/2}$-$\text{KLR}$}. Concerning the comparison between S-KLR and SVM, both approaches provide equivalent and substantially higher accuracy levels 
\textcolor{black}{than IVM and $\ell_{1/2}$-$\text{KLR}$}. 
It is worth emphasizing that S-KLR achieves a remarkable degree of sparsity compared \textcolor{black}{with standard KLR} and that the moderate sparsity difference in favor of the (intrinsically sparse) SVM is compensated by the more informative KLR probabilistic

\begin{center}
\rotatebox[origin=bl]{90}{
        \begin{minipage}[t]{1.55\textwidth}
        
            \begin{table}[H] 

            \caption{Comparative results in terms of \textcolor{black}{average testing accuracy and average percentage of data points involved in the model sparse representation ($\%$ data points)}  \textcolor{black}{obtained with} S-KLR, $\ell_{1/2}$-$\text{KLR}$, IVM and SVM. 
	For S-KLR, $\ell_{1/2}$-$\text{KLR}$ and SVM the additional columns ``sparsest of 3 most accurate" represent the best sparsity results over the three most accurate models in terms of accuracy. The standard deviation of the average accuracy is reported in parentheses.}\label{tab:results}
            \resizebox{1\textwidth}{!}{
            \textcolor{black}{
                \begin{tabular}{lcccccccccccccc}
\hline
\multicolumn{1}{c}{}            & \multicolumn{2}{c}{S-KLR}                       & \multicolumn{2}{c}{S-KLR}                  & \multicolumn{2}{c}{$\ell_{1/2}$-KLR}            & \multicolumn{2}{c}{$\ell_{1/2}$-KLR}        & \multicolumn{2}{c}{IVM}                     & \multicolumn{2}{c}{SVM}                         & \multicolumn{2}{c}{SVM}          \\
\multicolumn{1}{c}{}            & \multicolumn{2}{c}{sparsest of 3 most accurate} & \multicolumn{2}{c}{}                       & \multicolumn{2}{c}{sparsest of 3 most accurate} & \multicolumn{2}{c}{}                        & \multicolumn{2}{c}{}                        & \multicolumn{2}{c}{sparsest of 3 most accurate} & \multicolumn{2}{c}{}             \\ \cline{2-15} 
Dataset                         & accuracy         & ratio of selected             & accuracy      & ratio of selected           & accuracy        &ratio of selected              & accuracy      & ratio of selected            & accuracy      & ratio of selected            & accuracy         & ratio of selected            & accuracy      & ratio of selected \\ 
                         &          & data points             &       & data points           &         & data points              &       & data points            &      & data points            &          &  data points             &       &  data points \\\hline
\multicolumn{1}{l|}{banknote}   & 1                & \multicolumn{1}{c|}{0.06}    & 1             & \multicolumn{1}{c||}{0.162} & 0.835           & \multicolumn{1}{c|}{0.576}    & 0.935         & \multicolumn{1}{c||}{0.985}  & 0.882         & \multicolumn{1}{c||}{0.008}  & 1                & \multicolumn{1}{c|}{0.009}   & 1             & 0.009            \\
\multicolumn{1}{l|}{coil2000}   & 0.94             & \multicolumn{1}{c|}{0.155}   & 0.94          & \multicolumn{1}{c||}{0.265} & 0.94            & \multicolumn{1}{c|}{0.0002}   & 0.94          & \multicolumn{1}{c||}{0.0002} & 0.94          & \multicolumn{1}{c||}{0.0001} & 0.94             & \multicolumn{1}{c|}{0.119}   & 0.94          & 0.119            \\
\multicolumn{1}{l|}{diabetes}   & 0.77             & \multicolumn{1}{c|}{0.529}   & 0.767         & \multicolumn{1}{c||}{0.724} & 0.678           & \multicolumn{1}{c|}{0.171}    & 0.755         & \multicolumn{1}{c||}{0.747}  & 0.653         & \multicolumn{1}{c||}{0.002}  & 0.745            & \multicolumn{1}{c|}{0.49}    & 0.767         & 0.531            \\
\multicolumn{1}{l|}{ionosphere} & 0.946            & \multicolumn{1}{c|}{0.488}   & 0.946         & \multicolumn{1}{c||}{0.488} & 0.726           & \multicolumn{1}{c|}{0.011}    & 0.946         & \multicolumn{1}{c||}{0.304}  & 0.879         & \multicolumn{1}{c||}{0.221}  & 0.926            & \multicolumn{1}{c|}{0.444}   & 0.934         & 0.459            \\
\multicolumn{1}{l|}{magic}      & 0.857            & \multicolumn{1}{c|}{0.69}    & 0.857         & \multicolumn{1}{c||}{0.754} & 0.648           & \multicolumn{1}{c|}{0}        & 0.741         & \multicolumn{1}{c||}{0.989}  & 0.719         & \multicolumn{1}{c||}{0.0001} & 0.875            & \multicolumn{1}{c|}{0.293}   & 0.874         & 0.299            \\
\multicolumn{1}{l|}{monk2}      & 0.958            & \multicolumn{1}{c|}{0.292}   & 0.958         & \multicolumn{1}{c||}{0.297} & 0.528           & \multicolumn{1}{c|}{0}        & 0.937         & \multicolumn{1}{c||}{0.413}  & 0.833         & \multicolumn{1}{c||}{0.027}  & 0.974            & \multicolumn{1}{c|}{0.222}   & 0.976         & 0.286            \\
\multicolumn{1}{l|}{ring}       & 0.978            & \multicolumn{1}{c|}{0.085}   & 0.978         & \multicolumn{1}{c||}{0.113} & 0.505           & \multicolumn{1}{c|}{0.135}    & 0.744         & \multicolumn{1}{c||}{0.857}  & 0.763         & \multicolumn{1}{c||}{0.002}  & 0.974            & \multicolumn{1}{c|}{0.07}    & 0.978         & 0.09             \\
\multicolumn{1}{l|}{sonar}      & 0.866            & \multicolumn{1}{c|}{0.924}   & 0.856         & \multicolumn{1}{c||}{0.941} & 0.534           & \multicolumn{1}{c|}{0}        & 0.756         & \multicolumn{1}{c||}{0.464}  & 0.822         & \multicolumn{1}{c||}{0.347}  & 0.643            & \multicolumn{1}{c|}{0.924}   & 0.851         & 0.923            \\
\multicolumn{1}{l|}{spambase}   & 0.935            & \multicolumn{1}{c|}{0.246}   & 0.935         & \multicolumn{1}{c||}{0.518} & 0.675           & \multicolumn{1}{c|}{0.302}    & 0.804         & \multicolumn{1}{c||}{0.913}  & 0.902         & \multicolumn{1}{c||}{0.008}  & 0.931            & \multicolumn{1}{c|}{0.173}   & 0.935         & 0.206            \\
\multicolumn{1}{l|}{twonorm}    & 0.976            & \multicolumn{1}{c|}{0.075}   & 0.978         & \multicolumn{1}{c||}{0.141} & 0.487           & \multicolumn{1}{c|}{0}        & 0.976         & \multicolumn{1}{c||}{0.752}  & 0.977         & \multicolumn{1}{c||}{0.002}  & 0.975            & \multicolumn{1}{c|}{0.075}   & 0.977         & 0.101            \\
\multicolumn{1}{l|}{waveform}   & 0.91             & \multicolumn{1}{c|}{0.282}   & 0.91          & \multicolumn{1}{c||}{0.516} & 0.899           & \multicolumn{1}{c|}{0.568}    & 0.905         & \multicolumn{1}{c||}{0.922}  & 0.884         & \multicolumn{1}{c||}{0.005}  & 0.909            & \multicolumn{1}{c|}{0.287}   & 0.909         & 0.287            \\
\multicolumn{1}{l|}{wisconsin}  & 0.974            & \multicolumn{1}{c|}{0.145}   & 0.975         & \multicolumn{1}{c||}{0.143} & 0.805           & \multicolumn{1}{c|}{0.039}    & 0.96          & \multicolumn{1}{c||}{0.707}  & 0.936         & \multicolumn{1}{c||}{0.052}  & 0.965            & \multicolumn{1}{c|}{0.115}   & 0.975         & 0.147            \\ \hline
\multicolumn{1}{l|}{Average}    & 0.926 (0.066)    & \multicolumn{1}{c|}{0.331}   & 0.925 (0.067) & \multicolumn{1}{c|}{0.422} & 0.688 (0.157)   & \multicolumn{1}{c|}{0.15}     & 0.867 (0.097) & \multicolumn{1}{c|}{0.671}  & 0.849 (0.096) & \multicolumn{1}{c|}{0.056}  & 0.905 (0.106)    & \multicolumn{1}{c|}{0.268}   & 0.926 (0.067) & 0.288            \\ \hline
\end{tabular}
}
            }
            \end{table}      
        \end{minipage}
        }
\end{center}

\noindent information.

\textcolor{black}{In the second type of experiments, we first assess the improvement in model sparsity and then we evaluate the reduction in CPU training time.}

\textcolor{black}{As far as sparsity is concerned, we compare the proposed S-KLR approach with the original formulation \eqref{eq:k-dual} solved using a SMO-type algorithm based on the first-order working set selection procedure in \cite{keerthi2005fast}, denoted as KLR-1ord}.
\textcolor{black}{We also consider KLR-2ord which extends KLR-1ord by adopting the same second-order WSS procedure devised for S-KLR}
\footnote{\textcolor{black}{For a better comparison with KLR-2ord and S-KLR, the KLR-1ord implementation slightly differs from the method in \cite{keerthi2005fast}. To improve numerical stability, in \cite{keerthi2005fast} whenever a variable reaches an auxiliary bound it is fixed at that bound and only reconsidered in a second refinement phase. Although in KLR-1ord the auxiliary bounds are only used to constrain subproblem \eqref{eq:onedim_pr}, the difference with respect to the method in \cite{keerthi2005fast} is negligible due to the strict convexity of the original problem (both versions converge to the same unique solution).
}}. \textcolor{black}{Notice that KLR-2ord is a special case of S-KLR obtained by setting $\lambda = 0$, that is, S-KLR generalizes KLR-2ord with the sparsity inducing variant of the KLR formulation}. \textcolor{black}{It is worth pointing out that the comparison between S-KLR and KLR-2ord aims at verifying that the high level of sparsity achieved by S-KLR is essentially due to the $\rho$ term in the sparse KLR formulation \eqref{eq:2BM}, and not to the second-order WSS procedure. As mentioned in Section \ref{subsec:SMO}, the latter is devised for efficiency reasons.}

The results of the sparsity
experiments are shown in Table \ref{tab:first_results}. \textcolor{black}{In particular, for each dataset, we report the average testing accuracy of KLR-1ord and KLR-2ord, obtained by selecting the optimal hyperparameter $C$ on a validation set, as explained above. In the case of S-KLR, the reported accuracy corresponds to the average performance achieved by jointly tuning $C$ and $\lambda$.}
Comparing S-KLR with KLR-2ord, we observe that the sparsity term in S-KLR yield a substantial reduction in the number of data points that are actually used to determine the decision boundaries (from \textcolor{black}{$85.2\%$ to $42.2\%$}). The second-order WSS procedure in KLR-2ord seems to lead to a slight increase in the degree of sparsity with respect to KLR-1ord. \textcolor{black}{In terms of accuracy, both S-KLR and KLR-2ord yield a moderate improvement with respect to KLR-1ord.}  

\begin{table}[h]
	\caption{\textcolor{black}{Average testing accuracy and average percentage of data points involved in the model sparse representation ($\%$ data points)} \textcolor{black}{achieved by} KLR-1ord, KLR-2ord, and S-KLR. 
		KLR-1ord coincides with the method proposed in \cite{keerthi2005fast}. The standard deviation of the average accuracy is reported in parentheses. }\label{tab:first_results}
	\centering
	\resizebox{1\textwidth}{!}{
    \textcolor{black}{
		\begin{tabular}{lcccccc}
\hline
                                & \multicolumn{2}{c}{KLR 1-ord (\cite{keerthi2005fast})}              & \multicolumn{2}{c}{KLR 2-ord}              & \multicolumn{2}{c}{S-KLR}        \\ \cline{2-7} 
Dataset                         & accuracy      &  ratio of selected           & accuracy      & ratio of selected           & accuracy      &  ratio of selected \\
                         &       &  data points           &       &  data points           &       & data points \\\hline
\multicolumn{1}{l|}{banknote}   & 0.977         & \multicolumn{1}{c|}{1}     & 1             & \multicolumn{1}{c|}{0.694} & 1             & 0.162            \\
\multicolumn{1}{l|}{coil2000}   & 0.94          & \multicolumn{1}{c|}{0}     & 0.94          & \multicolumn{1}{c|}{0.155} & 0.94          & 0.265            \\
\multicolumn{1}{l|}{diabetes}   & 0.763         & \multicolumn{1}{c|}{1}     & 0.776         & \multicolumn{1}{c|}{1}     & 0.767         & 0.724            \\
\multicolumn{1}{l|}{ionosphere} & 0.943         & \multicolumn{1}{c|}{1}     & 0.94          & \multicolumn{1}{c|}{1}     & 0.946         & 0.488            \\
\multicolumn{1}{l|}{magic}      & 0.839         & \multicolumn{1}{c|}{1}     & 0.807         & \multicolumn{1}{c|}{0.69}  & 0.857         & 0.754            \\
\multicolumn{1}{l|}{monk2}      & 0.96          & \multicolumn{1}{c|}{1}     & 0.972         & \multicolumn{1}{c|}{0.96}  & 0.958         & 0.297            \\
\multicolumn{1}{l|}{ring}       & 0.949         & \multicolumn{1}{c|}{1}     & 0.978         & \multicolumn{1}{c|}{0.817} & 0.978         & 0.113            \\
\multicolumn{1}{l|}{sonar}      & 0.823         & \multicolumn{1}{c|}{1}     & 0.823         & \multicolumn{1}{c|}{1}     & 0.856         & 0.941            \\
\multicolumn{1}{l|}{spambase}   & 0.925         & \multicolumn{1}{c|}{1}     & 0.934         & \multicolumn{1}{c|}{0.95}  & 0.935         & 0.518            \\
\multicolumn{1}{l|}{twonorm}    & 0.978         & \multicolumn{1}{c|}{1}     & 0.979         & \multicolumn{1}{c|}{0.987} & 0.978         & 0.141            \\
\multicolumn{1}{l|}{waveform}   & 0.909         & \multicolumn{1}{c|}{1}     & 0.911         & \multicolumn{1}{c|}{0.998} & 0.91          & 0.516            \\
\multicolumn{1}{l|}{wisconsin}  & 0.952         & \multicolumn{1}{c|}{1}     & 0.951         & \multicolumn{1}{c|}{0.978} & 0.975         & 0.143            \\ \hline
Average                         & 0.913 (0.068) & \multicolumn{1}{c|}{0.917} & 0.918 (0.074) & \multicolumn{1}{c|}{0.852} & 0.925 (0.068) & 0.422            \\ \hline
\end{tabular}
}
	}
\end{table}
\textcolor{black}{Concerning CPU training time, Table \ref{tab:data_tempi} reports the training times in seconds required by the S-KLR using the second order WSS procedure (denoted as S-KLR-2ord) and a benchmarking version of S-KLR implementing the same first order WSS in \cite{keerthi2005fast} (denoted as S-KLR-1ord).} \textcolor{black}{For each dataset, the average CPU times over all the values of the grid search on the $C$ and $\lambda$ hyperparameters are indicated. In particular, the first two columns correspond to S-KLR-1ord and S-KLR-2ord, while the last column shows the ratio between the latter and the former CPU times.}

\textcolor{black}{The computational results show that S-KLR-2ord yields an average CPU time saving of about $18\%$. Except for spambase, where S-KLR-1ord is slightly faster (about $1\%$), S-KLR-2ord always leads to a CPU time reduction, often above the $20\%$ of time savings, with peaks of $36.6\%$ (ionosphere) and $38.2\%$ (wisconsin).} \textcolor{black}{It is worth emphasizing that such favorable results of S-KLR-2ord with respect to S-KLR-1ord are not obvious. Indeed, since the objective function of the sparse KLR formulation \eqref{eq:2BM} is not quadratic, the \textcolor{black}{second-order} information used to select the second WSS index may not be more reliable than the first-order one used in S-KLR-1ord. This clearly differs from the SVM case in which, due to the quadratic objective function, the information used in the second-order WSS is exact. }

\begin{table}[h]

	\centering
	\caption{\textcolor{black}{Average training time required by} S-KLR-1ord and S-KLR-2ord for all the 12 datasets.}\label{tab:data_tempi}
	\begin{tabular}{lccc}
		\hline
			& \multicolumn{2}{c}{CPU time (sec.)}    & \multicolumn{1}{c}{Ratio} \\ \cline{2-4} 
		Dataset                 & \multicolumn{1}{c}{S-KLR-1ord} & S-KLR-2ord                       & 2ord/1ord                     \\ \hline
		banknote                & 9.649                         & 7.559                    & 0.783                      \\
		coil2000                & 22.654                         & 16.633                    & 0.734                     \\
		diabetes                & 8.36                         & 7.068                     & 0.845                      \\
		ionosphere              & 10.955                       & 6.989                     & 0.638                     \\
		magic                   & 53.779                            & 52.841                   & 0.982                     \\
		monk2                   & 6.388                            & 4.548                     & 0.712                      \\
		ring                & 21.12                             & 20.011                    & 0.947                     \\
		sonar                   & 6.85                            & 5.204                     & 0.759                     \\
		spambase                & 16.282                         & 16.525                    & 1.015                     \\
		twonorm                 & 25.21                          & 24.73                    & 0.981                     \\
		waveform                 & 16.658                         & 14.077                    & 0.845                     \\
		wisconsin & 8.839                        & 5.464 & 0.618 \\ \hline
  Average & 17.23 (13.2)                        & 15.13 (13.6) & 0.821 \\ \hline
	\end{tabular}
\end{table}

\section{\textcolor{black}{Concluding remarks}}
\label{sec:conclusion}

\textcolor{black}{We revisited the problem of training sparse binary KLR so as to improve the balance between model sparsity and testing accuracy with respect to the available approaches. 
In particular, we presented} \textcolor{black}{a variant of the strictly convex} \textcolor{black}{KLR formulation capable of inducing sparsity and a SMO-type training algorithm to exactly solve its dual, which exploits second-order information and admits asymptotic convergence guarantees.}


\textcolor{black}{S-KLR was tested on 12 benchmark datasets and compared with the sparse KLR heuristic IVM, the regularization approach $\ell_{1/2}$-$\text{KLR}$, and the popular sparse kernel model SVM. The numerical results indicate} that S-KLR and SVM achieve a better trade-off between testing accuracy and sparsity than IVM \textcolor{black}{and $\ell_{1/2}$-$\text{KLR}$}. S-KLR and SVM obtain both similar high accuracy levels and good sparsity levels. The sensibly lower sparsity of S-KLR with respect to SVM is compensated by the 
\textcolor{black}{informative class probability estimates provided by the former model.} \textcolor{black}{As far as training times are concerned,  the proposed second order WSS procedure in our SMO algorithm turns out to have a significant impact.}
\textcolor{black}{From a theoretical point of view, we investigated the relationship between primal variable $\rho$ enforcing sparsity in the primal formulation and the dual hyperparameter $\lambda$. Moreover, we derived upper bounds on the $\lambda$ hyperparameter beyond which a subset of all the dual variables $\alpha_i$ of all of them reach $C$ up to a small tolerance required for numerical stability.} Empirical evidence \textcolor{black}{suggests that S-KLR does not require a time consuming grid search procedure involving both hyperparameters $C$ and $\lambda$. Indeed, for the considered datasets,} the value of the sparsity hyperparameter $\lambda$ can be set \textcolor{black}{according to a simple} rule without compromising the performance. 


\textcolor{black}{Interesting research directions include: the adoption of algorithmic strategies like kernel caching, low-rank kernel approximations, and parallel training, to improve the SMO-type algorithm in order to scale to larger datasets; the development of a KLR variant which induces feature sparsity (only a subset of features are actually used); the extension of the S-KLR method to multiclass classification.}
\textcolor{black}{Finally, we could also try to gain further theoretical insight into the level of sparsity obtained by the S-KLR model via the proposed formulation. }





\bibliographystyle{elsarticle-num} 
\bibliography{mybibliography.bib}

\appendix

\section{\textcolor{black}{Considerations on dual variables as the regularization parameter}}\label{sec:bounds_lambda}
\textcolor{black}{For the sake of notation we define $I_+=\{i \in I :y_i=+1\}$, $I_-=\{i \in I :y_i=-1\}$, $n_+=|I_+|$, $n_-=|I_-|$, and
$Q_{ij}=y_i y_j K_{ij}.$}\par
\textcolor{black}{Recalling that we assume matrix $Q$ to be positive semidefinite, for convenience we report the convex formulation \eqref{eq:regularized}:
\begin{align*}
&\min\ \; f_\lambda(\alpha)
=\frac{1}{2}\alpha^\top Q\alpha
+ C\sum_{i=1}^n G\!\Big(\frac{\alpha_i}{C}\Big)
- \lambda \sum_{i=1}^n \alpha_i \\ 
&\;\;\;\;\text{s.t.} \quad y^\top\alpha=0.
\end{align*}
We define
\[
h(\alpha)=\frac{1}{2}\alpha^\top Q\alpha + C\sum_{i=1}^n G\!\Big(\frac{\alpha_i}{C}\Big) \;\;\text{and}\;\; f_\lambda(\alpha)=h(\alpha)-\lambda\,\1^\top\alpha.
\]}

\textcolor{black}{It is easy to show that the function $h(\bm{\alpha})$ is bounded on $[0,C]^n$.}
\vspace{-15pt}

\textcolor{black}{\begin{lemma}\label{lem:bound}
There exists \(M<\infty\) such that \(|h(\alpha)|\le M\) for every \(\alpha\in[0,C]^n\).
\end{lemma}}

\noindent \textcolor{black}{\textit{Proof}
Since $Q$ is symmetric and postive semidefinite, on \([0,C]^n\) we have that \\ \(\alpha^\top Q\alpha\le \mbox{eig}_{max}(Q)\|\alpha\|^2\ \le \mbox{eig}_{max}(Q)\,nC^2\) where $\mbox{eig}_{max}(Q)$ is the maximum eigenvalue of $Q$. Moreover \(G(\delta)\in[-\log2,0]\) for \(\delta\in[0,1]\). Thus
\(|h(\alpha)|\le M\) with \( M=\tfrac{1}{2}\mbox{eig}_{max}(Q)\, nC^2 + C\,n\,\log 2\).
\qed}

\textcolor{black}{By considering formulation \eqref{eq:regularized}, the presence of $G(\cdot)$ in the objective function implies that the optimal solution is contained in the hypercube $\left[0,C\right]^n$, let us define this bounded version of the feasible region \(\mathcal F=\{\alpha\in[0,C]^n:\ y^\top\alpha=0\}\) and $m^*=\max\limits_{\alpha\in\mathcal F}\ \sum\limits_{i \in I} \alpha_i$.}
\vspace{-15pt}

\textcolor{black}{\begin{lemma}\label{lem:mstar}
Let \(n_{\text{min}}=\min\{n_+,n_-\}\). Then 
$$
m^* = 2Cn_{\text{min}}.
$$
\end{lemma}}
\textcolor{black}{\noindent \textit{Proof}
    Let define $S_+ = \sum\limits_{i \in I_+} \alpha_i$ and $S_- = \sum\limits_{i \in I_-} \alpha_i$, then $S_+ \leq C n_+$ and $S_- \leq C n_-$. From the equality constraints $\bm{\alpha}^T \bm{y}=0$ we have that $S_+ = S_- \leq C n_-$ ($S_- = S_+ \leq C n_+$), thus $S_+ = S_- \leq C \min\{n_+,n_-\} = C n_{\text{min}}$. Therefore $m^* \leq 2Cn_{\text{min}}$. To obtain equality, it is sufficient to set $\alpha_i = C$ for all data points belonging to the minority class and for  $n_{\text{min}}$ data points from the majority class, while setting $\alpha_i=0$ for all remaining data points. 
\qed}

\textcolor{black}{\begin{proposition}\label{prop:limit_optimal}
For any given $\lambda \in \mathbb{R}$, let \(\bm{\alpha}(\lambda)\) denote the minimizer of \eqref{eq:regularized}. Then $$ \lim\limits_{\lambda \rightarrow \infty}
\sum\limits_{i \in I}\alpha_i(\lambda)= m^*=2Cn_{\text{min}}.
$$
\end{proposition}}

\noindent \textcolor{black}{\textit{Proof}
For any $\varepsilon>0$, if $\alpha\in\mathcal F$ satisfies $\sum\limits_{i \in I}\alpha_i\le m^*-\varepsilon$, from Lemma \ref{lem:bound} we have
$$
f_\lambda(\alpha)=h(\alpha)-\lambda\,\sum\limits_{i \in I}\alpha_i
\ge -M - \lambda(m^*-\varepsilon).
$$
From Lemma \ref{lem:mstar} we know that there exists $\bar\alpha\in\mathcal F$ with $\sum\limits_{i \in I}\bar{\alpha}_i=m^*$  and thus
$f_\lambda(\bar\alpha)\le M - \lambda m^*$.
For $\lambda>2M/\varepsilon$ we have $f_\lambda(\alpha)>f_\lambda(\bar\alpha)$.
Therefore, for sufficiently large values of $\lambda$, any minimizer $\alpha(\lambda)$ must satisfy
$\sum\limits_{i \in I}\alpha(\lambda)\ge m^*-\varepsilon$. Since this reasoning can be applied for any arbitrary small $\varepsilon$, we have:} 
\vspace{-5pt}

\textcolor{black}{$$ \lim\limits_{\lambda \rightarrow \infty}
\sum\limits_{i \in I}\alpha_i(\lambda)=  m^*=2Cn_{\text{min}}.
$$}
\vspace{-5pt}

\textcolor{black}{Since the sequence of $\bm\alpha(\lambda)$ for growing $\lambda$ is defined on a compact set $\cal{F}$, it admits a limit point $\bm\alpha(\lambda)$ which, from the strict convexity of the objective function in \eqref{eq:regularized}, is the unique optimal solution.  
\qed}

\textcolor{black}{In the case of balanced classes ($n_+=n_-$) we have $n_{\text{min}} = \frac{n}{2}$
 and $m^*= Cn$. From Proposition \ref{prop:limit_optimal} follows:}

\vspace{-15pt}

\textcolor{black}{\begin{corollary}\label{cor:uno}
For any dataset with balanced classes ($n_+=n_-$) the optimal solution $\bm{\alpha}(\lambda)$ of formulation \eqref{eq:regularized} is such that 
$$
\lim\limits_{\lambda \rightarrow \infty}\bm{\alpha}(\lambda) = C\,\1.
$$
\end{corollary}}
\vspace{-15pt}

\textcolor{black}{\setcounter{lemma}{0}
\begin{lemma}\label{lem:unbound_alpha}
For any value of $\lambda \in \mathbb{R}$ the optimal solution of formulation \eqref{eq:regularized} $\bm{\alpha}(\lambda)$ is such that $$0 < \alpha_i(\lambda) < C \quad \forall i \in I.$$
\end{lemma}}

\noindent \textcolor{black}{\textit{Proof}
Let us proceed by contradiction. Suppose that $\alpha_k(\lambda)=0$ for some index $k$.
The two sets
$$
\mathcal R = \{i:\ y_i=y_k,\ \alpha_i(\lambda)>0\} \;\; \text{and} \;\;
\mathcal S = \{i:\ y_i=-y_k,\ \alpha_i(\lambda)<C\}$$
cannot both be empty. Indeed, if they were, we would have that}

\textcolor{black}{$$\bm{y}^T\bm{\alpha}(\lambda) = \sum_{i:\,y_i=y_k}\!\!\! \alpha_i(\lambda)\,y_k
+ \sum_{i:\,y_i=-y_k}\!\!\! \alpha_i(\lambda)\,(-y_k)
= 0 - C\,y_k\,|\{i:\,y_i=-y_k\}|\ \neq\ 0,$$}

\noindent \textcolor{black}{thus $\bm{\alpha}(\lambda) \notin \cal{F}$. Therefore, we have either (i) at least one $m \in \cal{R}$, or (ii) at least one $i \in \cal{S}$.}


\textcolor{black}{Let us consider case i). Then, we can define a direction $\bm{d}$ with $d_k = 1$, $d_m = -1$, and $d_j = 0 \; \forall j \neq m,k$ such that $\bm{y}^T \bm{d} = 0$ and $\bm{\alpha}(\lambda) + \tau \bm{d} \in \cal{F}$ for any $\tau \in (0,\bar\tau]$ where $\bar\tau = C-\alpha_k(\lambda)$, hence $\bm d$ is a feasible direction at $\bm{\alpha}(\lambda)$. It is easy to see that $\nabla f_k(\bm\alpha(\lambda)) d_k<0$. Indeed, while the derivatives of $\frac{1}{2} \| \bm{\w}(\bm{\alpha})\|^2$ and $- \lambda \sum_i \alpha_i$ are bounded over $\cal{F}$, function $G$ is equal to $-\infty$ at zero (recall $\bm\alpha_k=0$). This contradicts the optimality condition $\nabla f(\bm{\alpha}(\lambda))^T \bm{d} \geq 0$.}\par
\textcolor{black}{Concerning case ii), we can follow the same argument of case i) (replacing $\cal{R}$ with $\cal{S}$ and considering $d_m=1$) to obtain the same contradiction. Therefore, for any $\lambda \in \mathbb{R}$ we can guarantee that  $\bm{\alpha}(\lambda) > \bm{0}$.}\par 
\textcolor{black}{We omit the proof of $\bm{\alpha}(\lambda) < C$ that can be obtained with the same arguments of the previous cases i) and ii) related to $\bm\alpha(\lambda)>0$ by simply inverting the orientation of the feasible direction $\bm d$.
\qed}

\textcolor{black}{We now recall the statement of Proposition \ref{prop:final_bound} (in Section \ref{sec:useful_prop}) and report the proof.}

\textcolor{black}{\setcounter{proposition}{0}
\begin{proposition}
Assume without loss of generality that $I_-$ is the minority class. Let $\bm\alpha(0)$ denote the optimal solution of formulation \eqref{eq:regularized} with $\lambda = 0$. Consider any positive tolerance $\gamma$ such that $\gamma< C-\max\limits_{i \in I_{-}}\alpha_i(0)$, and define $\bar{C} = C - \gamma$. For any \begin{equation}\label{eq:ubapp}
    \lambda \geq \max\limits_{i \in I_-} \sum\limits_{j \in I} (\bar{C}-\alpha_j(0)) y_j K_{ji},
\end{equation}
the optimal solution ${\bm\alpha}(\lambda)$ of formulation \eqref{eq:regularized} satisfies $\alpha_i(\lambda) \geq \bar{C} \;\, \forall i \in I_-$. 
\end{proposition}}

\noindent \textcolor{black}{\textit{Proof} From Proposition \ref{prop:limit_optimal}, we know that as $\lambda$ increases, $\alpha_i(\lambda) \; \forall i \in I_-$ tends to $C$. Therefore, there exists a sufficiently large $\bar \lambda$ for which for any $\lambda\geq \bar \lambda$ we have $\alpha_i(\lambda)>\alpha_i(0) \quad \forall i \in I_{-}$. For any $\lambda\geq \bar \lambda$, recalling the monotone relationship between dual $\alpha_i$ and primal $\xi_i$ (see \eqref{eq:oc_xi}), we have that $\xi_i(\lambda)>\xi_i(0)$, hence from  \eqref{eq:sparse-primal-con}}

\textcolor{black}{\begin{equation}\label{eq:proof_1}
    \rho(\lambda) - y_i (\bm{\omega}(\lambda)\cdot \bm{z}_i-b(\lambda)) \geq -y_i(\bm{\omega}(0)\cdot \bm{z}_i-b(0))\qquad \forall i \in I_-.
\end{equation}}

\textcolor{black}{Omitting the variable $b$ for convenience\footnote{\textcolor{black}{The intercept can be absorbed into the kernel by augmenting each input with a constant feature (or equivalently adjusting the kernel), so the explicit intercept term may be omitted without loss of generality.}} and using equation \eqref{eq:oc_w} into \eqref{eq:proof_1}, we obtain} 

\textcolor{black}{$$\rho(\lambda) > y_i \sum\limits_{j \in I} (\alpha_j(\lambda) - \alpha_j(0)) y_j K_{ij} \qquad \forall i \in I_{-}.$$}

\textcolor{black}{Thus, recalling that $\rho(\lambda)=\lambda$, to obtain $\alpha_i(\lambda)\geq\bar C \quad \forall i \in I_{-}$ it is sufficient to set} 

\textcolor{black}{$$\lambda \geq \max\limits_{i \in I_-} \sum\limits_{j \in I} (\bar{C}-\alpha_j(0))y_j K_{ij}.$$
\qed}

\section{Asymptotic convergence analysis} \label{subsec:prop_app}

In this section we provide the convergence analysis leading to the result of Proposition \ref{pr:pr1} stated in Section \ref{sec:dec}.

We start with a technical Lemma concerning the convergence of ``shifted" sub-sequences of convergent sub-sequences of any sequence $\{\bm{\alpha}^{k}\}$, $k \in \mathbb{N}$, made up of feasible solutions for problem \eqref{eq:2BM}.
\setcounter{lemma}{3}
\begin{lemma}
	\label{le:lemma2}
	If a sub-sequence $\{\bm{\alpha}^k\}$, $k \in {\cal K}$ converges to $\bm{\bar{\alpha}}$, then for any given positive integer $s$, the sub-sequence $\{\bm{\alpha}^{k+s}\}$, $k \in {\cal K}$, converges to $\bm{\bar{\alpha}}$ as well.
\end{lemma}
\noindent \textit{Proof} First, we verify that the inequality
\begin{equation}
    \label{eq:suff_dec}
    f(\bm{\alpha}^{k}) - f(\bm{\alpha}^{k+1}) \geq \frac{2}{C} \|\bm{\alpha}^{k+1} - \bm{\alpha}^{k} \|^2
\end{equation}

\noindent holds for every iteration index $k \geq 0.$ This inequality guarantees the sufficient decrease of the objective function with respect to the magnitude of the step.

Assuming \textcolor{black}{that} $t^*$ \textcolor{black}{denotes} the optimal solution of the one-dimensional subproblem \eqref{eq:onedim_pr}, we can express the \textcolor{mygreen}{second-order} Taylor expansion of $\phi(t)$ around $t^*$ \textcolor{black}{as}:
	\begin{equation}\label{eq:second_order}
		\phi(t) = \phi(t^*) + \frac{1}{2}\phi^{''}(\tilde{t})(t-t^*)^2,
	\end{equation}
	with $\tilde{t}$ such that $t \leq \tilde{t} \leq t^*$. By considering the \textcolor{mygreen}{second-order} derivative of $\phi(t)$:
	
	{\footnotesize
		$$\phi^{''}(t) = Ker(\bm{x}_{i^*},\bm{x}_{i^*})+Ker(\bm{x}_{j^*},\bm{x}_{j^*}) - 2Ker(\bm{x}_{i^*},\bm{x}_{j^*}) + \frac{1}{C}\left[  G^{''}(\frac{\tilde{\alpha}_{i^*}(t)}{C})+ G^{''}(\frac{\tilde{\alpha}_{j^*}(t)}{C})\right]$$}
  
  \noindent and the fact that $G^{''}(\cdot) \geq 4$ and the sum of the first three terms (involving $Ker(\cdot,\cdot)$) is greater or equal to zero, we have $\phi^{''}(t) \geq \frac{8}{C} $. From \eqref{eq:second_order}, for $t=0$ we have:
\begin{equation*}
  f(\bm{\alpha}^{k}) - f(\bm{\alpha}^{k+1}) = \phi(0) - \phi(t^*) \geq \frac{4}{C}(t^*)^2 = \frac{2}{C} \|\bm{\alpha}^{k+1} - \bm{\alpha}^{k} \|^2,  
\end{equation*}

\noindent which amounts to the sufficient decrease inequality.

\textcolor{black}{To prove the lemma, we use the previous inequality \eqref{eq:suff_dec} and the fact that the sequence $\{f(\bm{\alpha}^k)\}$ is not only decreasing but bounded, since the objective function \eqref{eq:final} is continuous over a compact set}.

Considering any sub-sequence $\{\bm{\alpha}^k\}$, $k \in {\cal K}$, which converges to $\bm{\bar{\alpha}}$, for the sub-sequence $\{\bm{\alpha}^{k+1}\}$, $k \in {\cal K}$, we have
	\begin{equation*} 
		\begin{split}
			\lim\limits_{k \in {\cal K}, k \rightarrow \infty} \|\bm{\alpha}^{k+1} - \bm{\bar{\alpha}} \| \leq \lim\limits_{k \in {\cal K}, k \rightarrow \infty} \Big(\|\bm{\alpha}^{k+1} - \bm{\alpha}^k \| + \|\bm{\alpha}^k  - \bm{\bar{\alpha}} \| \Big) \\ \leq \lim\limits_{k \in {\cal K}, k \rightarrow \infty} \Bigg(\sqrt{\frac{2}{C} \Big( f(\bm{\alpha}^{k}) - f(\bm{\alpha}^{k+1})} \Big) + \|\bm{\alpha}^k - \bm{\bar{\alpha}} \| \Bigg) = 0. 
		\end{split}
	\end{equation*}
	\noindent Thus, the shifted sub-sequence   $\{\bm{\alpha}^{k+1}\}$, $k \in {\cal K}$, converges to $\bar{\bm{\alpha}}$. Applying the same argument to the sub-sequence $\{\bm{\alpha}^{k+1}\}$, $k \in {\cal K}$, we can also prove that $\lim\limits_{k \in {\cal K}, k \rightarrow \infty} \bm{\alpha}^{k+2} = \bar{\bm{\alpha}}$. By induction, we have that $\lim\limits_{k \in {\cal K}, k \rightarrow \infty} \bm{\alpha}^{k+s} = \bar{\bm{\alpha}}$ for any given $s$.
	\qed

We now verify that the WSS is a special case of WSS with constant factor violating pair (WSS-CF) discussed in \cite{chen2006study}\footnote{Notice that in \cite{chen2006study} WSS-CF is referred to as WSS2.}.
In the WSS-CF procedure reported below, we consider only pairs $i \in I_{\text{up}}(\bm{\alpha})$ and $j \in I_{\text{low}}(\bm{\alpha})$ such that the corresponding violation of the optimality condition \eqref{eq:opt_cond} is at least as large as the image of a \textcolor{black}{contraction function} $h$\footnote{\textcolor{black}{As mentioned in \cite{chen2006study}, $h: \mathbb{R}\rightarrow \mathbb{R}$ can be any function that is strictly increasing for $x \geq 0$, satisfying the conditions $h(x) \leq x$ for all $x \geq 0$, and $h(0)= 0$.}} applied to the violation of the maximal violating pair, namely:
\begin{equation*}\label{eq:max-violation}
	 -y_{i^{MVP}} \nabla f(\bm{\alpha})_{i^{MVP}} -(-y_{j^{MVP}} \nabla f(\bm{\alpha})_{j^{MVP}}). 
\end{equation*}
\par 
\medskip

\noindent WSS-CF\\
\hrule
\medskip
\begin{itemize}
	\item[1)] Consider a fixed $0 < \theta \leq 1$ for all iterations. 
	\item[2)] Select any pair $i^* \in I_{\text{up}}(\bm{\alpha})$ and $j^* \in I_{\text{low}}(\bm{\alpha})$ such that 
	\begin{equation}\label{eq:ws_condition}\footnotesize
		-y_{i^*} \nabla f(\bm{\alpha})_{i^*} - ( -y_{j^*} \nabla f(\bm{\alpha})_{j^*}) \geq \theta(-y_{i^{MVP}} \nabla f(\bm{\alpha})_{i^{MVP}} -   (-y_{j^{MVP}} \nabla f(\bm{\alpha})_{j^{MVP}}) ) > 0 
	\end{equation}
	\item[3)] Return the pair $\{i^*,j^*\}$
\end{itemize}
\hrule
\bigskip

It is easy to verify that, if the $N \times N$ kernel matrix $K$ is positive semidefinite, WSS is a special case of WSS-CF. Indeed, consider the pair $\{i^{*},j^{*}\}$ selected by WSS where $i^* = i^{MVP}$. At each iteration $k$, by definition of WSS, the current feasible solution  $\bm\alpha^k$ satisfies
{\footnotesize$$ -\frac{(-y_{i^*} \nabla f(\bm{\alpha}^{k})_{i^*} - (- y_{j^*} \nabla f(\bm{\alpha}^{k})_{j^*}))^2}{q_{i^{*}j^{*}}} \leq -\frac{(-y_{i^{MVP}} \nabla f(\bm{\alpha}^{k})_{i^{MVP}} -(- y_{j^{MVP}} \nabla f(\bm{\alpha}^{k})_{j^{MVP}}))^2}{q_{i^{MVP}j^{MVP}}} $$} 

\noindent with $ q_{i^{*}j^{*}}$ and $q_{i^{MVP}j^{MVP}}$ positive.
Since the feasible set of \eqref{eq:2BM} is compact, for any pair $\{i,j\}$ the function $q_{ij}=Ker(\bm{x}_{i},\bm{x}_{i})+Ker(\bm{x}_{j},\bm{x}_{j}) - 2Ker(\bm{x}_{i},\bm{x}_{j}) + \frac{C}{\alpha_{i}(C-\alpha_{i})}+ \frac{C}{\alpha_{j}(C-\alpha_{j})}$ is continuous in $\bm{\alpha} $ and admits a maximum and a minimum over the feasible region of \eqref{eq:2BM}. Taking the minimum and the maximum of $q_{ij}$ over all the pairs of indices $\{i,j\}$ and all the feasible $\bm{\alpha}$, we have the following inequality:


$$(-y_{i^*} \nabla f(\bm{\alpha}^{k})_{i^*} -(-y_{j^*} \nabla f(\bm{\alpha}^{k})_{j^*}) )\geq \sqrt{\frac{\min\limits_{ij}\min\limits_{\bm{\alpha}}q_{ij}}{\max\limits_{ij}\max\limits_{\bm{\alpha}}q_{ij}}} (-y_{i^{MVP}} \nabla f(\bm{\alpha}^{k})_{i^{MVP}} -(- y_{j^{MVP}} \nabla f(\bm{\alpha}^{k})_{j^{MVP}})). $$
Thus, WSS is a special case 
of WSS-CF with $\theta = \sqrt{\frac{\min\limits_{ij}\min\limits_{\bm{\alpha}}q_{ij}}{\max\limits_{ij}\max\limits_{\bm{\alpha}}q_{ij}}}$.

Now we are ready to prove Proposition \ref{pr:pr1} after recalling it.

\setcounter{proposition}{1}
\begin{proposition}
	Assume \textcolor{black}{that} the kernel matrix $K$ is positive semidefinite. Let $\{\alpha^{k}\}$ be the infinite sequence generated by the SMO-type method Algorithm \ref{alg:smo}. Then $\{\alpha^k\}$ globally converges to the unique optimal solution of problem \eqref{eq:2BM}. 
\end{proposition}
\noindent \textit{Proof} First, since $K$ is positive semidefinite it is easy to verify that for any given pair $\{i,j\}$ selected in the WSS procedure, the corresponding one-dimensional subproblem \eqref{eq:onedim_pr} is strictly convex. Indeed,
	{\footnotesize 
		$$\phi^{''}(t) = Ker(\bm{x}_{i},\bm{x}_{i})+Ker(\bm{x}_{j},\bm{x}_{j}) - 2Ker(\bm{x}_{i},\bm{x}_{j}) + \frac{1}{C}\left[  G^{''}(\frac{\tilde{\alpha}_{i}(t)}{C})+ G^{''}(\frac{\tilde{\alpha}_{j}(t)}{C})\right] > 0.$$
	}
 
\noindent Since the feasible set of \eqref{eq:2BM} is compact, $\{\bm{\alpha}^k\}$ admits a convergent sub-sequence. Suppose that $\bar{\bm{\alpha}}$ is the limit point of a convergent sub-sequence of $\{\bm{\alpha}^k\}$. If $\bar{\bm{\alpha}}$ is not a stationary point of \eqref{eq:2BM} (the optimality conditions \eqref{eq:opt_cond} are not satisfied) there exists a pair $\{\bar{i},\bar{j}\}$ with
	\begin{equation}
		\label{eq:violp}
		\bar{i} = \text{arg}\max_{s \in I_{\text{up}}(\bar{\bm{\alpha}})} -y_s \nabla f(\bar{\bm{\alpha}})_s\; \text{and }\; \bar{j}= \text{arg}\min_{s \in I_{\text{low}}(\bar{\bm{\alpha}})} -y_s \nabla f(\bar{\bm{\alpha}})_s,
	\end{equation}
	whose maximal violation is denoted as  
	
	\begin{equation}
		\label{eq:delta}
		\Delta = -y_{\bar{i}} \nabla f(\bar{\bm{\alpha}})_{\bar{i}} + y_{\bar{j}} \nabla f(\bar{\bm{\alpha}})_{\bar{j}} > 0.
	\end{equation} 
	
	We further define the smallest (non zero) absolute value of the difference between any possible pair of terms $-y_{\cdot}\nabla f(\bar{\bm{\alpha}})_{\cdot}$ as:
	\begin{equation}
		\label{eq:deltaprime}
		\Delta'=\min\limits_{t,s \in I} \Bigl\{|-y_{t}\nabla f(\bar{\bm{\alpha}})_{t}+y_{s}\nabla f(\bar{\bm{\alpha}})_{s}| \Bigr|-y_{t}\nabla f(\bar{\bm{\alpha}})_{t}\neq-y_{s}\nabla f(\bar{\bm{\alpha}})_{s}\Bigr\}>0.
	\end{equation}

 Given Lemma \ref{le:lemma2}, the continuity of $\nabla f(\bm{\alpha})$, and the fact that $\textcolor{black}{\theta\frac{\Delta'}{2}>0}$ due to \eqref{eq:deltaprime}, we can verify the claims \eqref{eq:dim_1}-\eqref{eq:dim_6}. For any given $r$, there exists $\bar{k}\in \mathcal{K}$ such that for all $k\in \mathcal{K},$ with $ k\geq\bar{k}$, the following hold:
\vspace{-0.55cm}

\begin{align}
		&\text{for the pair of indices } \{\bar{i},\bar{j}\}\, \text{as in } \eqref{eq:violp}\text{ and for } u=0, \ldots, r,\, \text{we have }\nonumber \\
  & -y_{\bar{i}}\nabla f(\bm{\alpha}^{k+u})_{\bar{i}}+y_{\bar{j}}\nabla f(\bm{\alpha}^{k+u})_{\bar{j}}>\Delta',\label{eq:dim_1}\\
		&\text{if }\, i\in I_{\text{up}}(\bar{\bm{\alpha}}) , \text{ then}\; i\in I_{\text{up}}(\bm{\alpha}^{k}) , .. . , i\in I_{\text{up}}(\bm{\alpha}^{k+r}),\label{eq:dim_2}\\
		&\text{if }\, i\in I_{\text{low}}(\bar{\bm{\alpha}}) , \text{ then}\; i\in I_{\text{low}}(\bm{\alpha}^{k}) , .. . , i\in I_{\text{low}}(\bm{\alpha}^{k+r}),\label{eq:dim_3}\\
		&\text{if for a given pair } \{i,j\} \text{ we have} -y_{i}\nabla f(\bar{\bm{\alpha}})_{i}+y_{j}\nabla f(\bar{\bm{\alpha}})_{j}> 0,\,\text{then for}\nonumber \\ & u=0, . . . ,r, \text{ we have }-y_{i} \nabla f(\bm{\alpha}^{k+u})_{i}+y_{j}\nabla f(\bm{\alpha}^{k+u})_{j}>\Delta',\label{eq:dim_4}\\
		&\text{if for a given pair } \{i,j\}\,\text{ we have } -y_{i}\nabla f(\bar{\bm{\alpha}})_{i}=-y_{j}\nabla f(\bar{\bm{\alpha}})_{j},\,\text{then for}\nonumber \\ & u=0, \ldots, r,\,\text{we have } |-y_{i}\nabla f(\bm{\alpha}^{k+u})_{i}+y_{j}\nabla f(\bm{\alpha}^{k+u})_{j}|<\textcolor{black}{\theta\Delta'},\label{eq:dim_5}\\
		&\text{if } \{i,j\} \text{ is the working set at the } (k+u)\text{-}\mathrm{th}\; \text{with }0\leq u\leq r-1 \text{ and} \nonumber \\
        & -y_{i}\nabla f(\bar{\bm{\alpha}})_{i} + y_{j}\nabla f(\bar{\bm{\alpha}})_{j}>0\;\text{then}\; i\not\in I_{\text{up}}(\bm{\alpha}^{k+u+1}) \;\text{or}\; j\not\in I_{\text{low}}(\bm{\alpha}^{k+u+1}).\label{eq:dim_6}
	\end{align}

	Let us verify the claim \eqref{eq:dim_1} which states that, for any iteration $k \geq \bar{k}$ the violation of the pair $\{\bar{i},\bar{j}\}$ for $\bm{\alpha}^{k+u}$ is larger than $ \Delta'$, for $u=0,...,r$. Lemma \ref{le:lemma2} implies that the sequences $\{\bm{\alpha}^{k}\}, \{\bm{\alpha}^{k+1}\}, ...,$ $ \{\bm{\alpha}^{k+r}\},$ $ k\in \mathcal{K}$, all converge to $\bar{\bm{\alpha}}$. From the continuity of $\nabla f(\bm{\alpha})$ and relation \eqref{eq:deltaprime}, for any given $u$ with $0\leq u\leq r$ and the corresponding sequence $\{\bm{\alpha}^{k+u}\}, k\in \mathcal{K}$, there exists $k_{u}$ such that \eqref{eq:dim_1} holds for all $k\geq k_{u},\, k\in \mathcal{K}$. Since $r$ is finite, by setting $\bar{k}$ to be the largest of these $k_{u}$ with $u=0,..., r,$ \eqref{eq:dim_1} is valid for all $u=0, \ldots, r$. 
    The derivations of \eqref{eq:dim_2}-\eqref{eq:dim_5} follow from similar arguments so they are omitted. 
 
 We now verify the claim \eqref{eq:dim_6} which states that, for any iteration $k \geq \bar{k}$, if at iteration $k + u$ with $0 \leq u \leq r - 1$ the working set pair $\{i,j\}$ has a positive violation associated to $\bar{\alpha}$, then at iteration $k+u+1$ either $i$ does not belong to $I_{\text{up}}(\bm{\alpha}^{k+u+1})$ or $j$ does not belong to  $I_{\text{low}}(\bm{\alpha}^{k+u+1})$. First, we define $\bm{\alpha}(t)$ as $\alpha_s(t)=\alpha^{k+u}_s$ with $s \neq i,j$ and $\alpha_i(t) =\alpha^{k+u}_i(t^*)$ and $\alpha_j(t)=\alpha^{k+u}_j(t^*)$, where $\alpha^{k+u}_i(t^*)\,\text{and}\, \alpha^{k+u}_j(t^*)$ are the optimal values of, respectively, $\alpha_i$ and $\alpha_j$ generated by solving the one-dimensional problem \eqref{eq:onedim_pr} at $k+u$ iteration. Similar to the optimality condition \eqref{eq:opt_cond} for problem \eqref{eq:2BM}, since $\bm{\alpha}(t)$ is stationary for the subproblem  then 
	\begin{equation}\label{eq:cond_opt_proof}
		\max\limits_{s \in I_{\text{up}}(\bm{\alpha}(t)): s \in \{i,j\}}-y_{s}\nabla f(\bm{\alpha}(t))_{s} \leq \min\limits_{s \in I_{\text{low}}(\bm{\alpha}(t)): s \in \{i,j\}}-y_{s}\nabla f(\bm{\alpha}(t))_{s}. 
	\end{equation}
	
	\noindent By noticing that $\bm{\alpha}(t) = \bm{\alpha}^{k+u+1}$, if
	
	$$i\in I_{\text{up}}(\bm{\alpha}^{k+u+1})\; \text{and}\; j \in I_{\text{low}}(\bm{\alpha}^{k+u+1}),$$
	
	\noindent then from \eqref{eq:cond_opt_proof} we have
	\begin{equation}
		-y_{i}\nabla f(\bm{\alpha}^{k+u+1})_{i} +y_{j}\nabla f(\alpha^{k+u+1})_{j} \leq 0.
	\end{equation}
	However, assuming that $-y_{i}\nabla f(\bar{\bm{\alpha}})_{i}+ y_{j}\nabla f(\bar{\bm{\alpha}})_{j} > 0$, claim \eqref{eq:dim_4} for $\bm{\alpha}^{k+u+1}$  implies that $-y_{i} \nabla f(\bm{\alpha}^{k+u+1})_{i}+y_{j}\nabla f(\bm{\alpha}^{k+u+1})_{j}>\Delta'$, so there is a contradiction.
	
	For ease of proof presentation, the indices of $\bar{\bm{\alpha}}$ are reordered as follows:
	
	\begin{equation}
		\label{eq:ordinamento}
		-y_{1}\nabla f(\bar{\bm{\alpha}})_{1}\leq ... \leq-y_{N}\nabla f(\bar{\bm{\alpha}})_{N}.
	\end{equation}

	\noindent We also introduce
	\begin{equation}
		S_{1}(k) =\sum\{i|i\in I_{\text{up}}(\bm{\alpha}^{k})\}\; \text{and}\; S_{2}(k) =\sum\{N-i|i\in I_{\text{low}}(\bm{\alpha}^{k})\}
	\end{equation}
	Clearly,
	\begin{equation}
		\label{eq:nocontr}
		N\leq S_{1}(k)+S_{2}(k)\leq N(N-1).
	\end{equation}

	If $\{i,j\}$ is selected at the $(k+u)$-$\mathrm{th}$ iteration (with $u=0,. ..,r$), then we must have that
	\begin{equation}
		\label{eq:geij}
		-y_{i}\nabla f(\bar{\bm{\alpha}})_{i} + y_{j}\nabla f(\bar{\bm{\alpha}})_{j}>0. 
	\end{equation}

\noindent Indeed, using an adaptation of claim \eqref{eq:dim_4} where we assume that $-y_{i}\nabla f(\bar{\bm{\alpha}})_{i}+y_{j}\nabla f(\bar{\bm{\alpha}})_{j}<0$, we obtain $-y_{i}\nabla f(\bm{\alpha}^{k+u})_{i}+y_{j}\nabla f(\bm{\alpha}^{k+u})_{j}<0$ which violates WSS condition \eqref{eq:ws_condition}. Moreover, if $-y_{i}\nabla f(\bar{\bm{\alpha}})_{i}+y_{j}\nabla f(\bar{\bm{\alpha}})_{j}=0$ then
\begin{align}\label{eq:ineq_proof}
		-y_{i}\nabla f(\bm{\alpha}^{k+u})_{i}+y_{j}\nabla f(\bm{\alpha}^{k+u})_{j} < \theta(\Delta')
  <\theta(-y_{\bar{i}}\nabla f(\bm{\alpha}^{k+u})_{\bar{i}}+y_{\bar{j}}\nabla f(\bm{\alpha}^{k+u})_{\bar{j}}) \nonumber \\
		\leq \theta(\max\limits_{s \in I_{\text{up}}(\bm{\alpha}^{k+u})}-y_{s}\nabla f(\bm{\alpha}^{k+u})_{s}-\min\limits_{s \in I_{\text{low}}(\bm{\alpha}^{k+u})}-y_{s}\nabla f(\bm{\alpha}^{k+u})_{s}).
	\end{align}
    Starting form the left, the first strict inequality is due to \eqref{eq:dim_5}, and the second strict inequality is due to \eqref{eq:dim_1}. The third inequality is implied by claims \eqref{eq:dim_2} and \eqref{eq:dim_3} where $\bar{i} \in I_{\text{up}}(\bar{\bm{\alpha}})$, and respectively $\,\bar{j}\in I_{\text{low}}(\bar{\bm{\alpha}})$. As a consequence, \eqref{eq:geij} is valid since \eqref{eq:ineq_proof} contradicts \eqref{eq:ws_condition}\footnote{Notice that in the last inequality in \eqref{eq:ineq_proof} we do not use the notation $i^{MVP}$ and $j^{MVP}$ because we want to explicitly refer to the $k+u$-th iterations.}. 
    
    To obtain the contradiction of \eqref{eq:violp} and \eqref{eq:delta} we exploit a counting procedure. Considering two consecutive iterations $k$-th and $k+1$-th, the \eqref{eq:geij} established above and \eqref{eq:dim_6} imply 
	$$i\not\in I_{\text{up}}(\bm{\alpha}^{k+1})\; \text{or}\;j\not\in I_{\text{low}}(\bm{\alpha}^{k+1}).$$

    Let us suppose that $i\not\in I_{\text{up}}(\bm{\alpha}^{k+1})$). Given \eqref{eq:dim_6} and \eqref{eq:geij}, from \eqref{eq:dim_2} we would have $i\not\in I_{\text{up}}(\bar{\bm{\alpha}})$ and hence $i\in I_{\text{low}}(\bar{\bm{\alpha}}).$ Due to \eqref{eq:dim_3} and the selection rule \eqref{eq:ws_condition}, we have $i\in I_{\text{low}}(\bm{\alpha}^{k})\cap I_{\text{up}}(\bm{\alpha}^{k}).$ Thus
	
	$$i\in I_{\text{low}}(\bm{\alpha}^{k})\cap I_{\text{up}}(\bm{\alpha}^{k})\;\text{and} \; i\not\in I_{\text{up}}(\bm{\alpha}^{k+1}).$$
	
\noindent	Since $j\in I_{\text{low}}(\bm{\alpha}^{k})$, we would have
	\begin{equation}
		\label{eq:s_k}
		S_{1}(k+1)\leq S_{1}(k)-i+j\leq S_{1}(k)-1\;\; \text{and}\;\; S_{2}(k+1)\leq S_{2}(k),
	\end{equation}

	\noindent where $-i+j\leq-1$ comes from \eqref{eq:ordinamento}. 
 
 Now suppose that $j\not\in I_{\text{low}}(\bm{\alpha}^{k+1})$. With similar arguments we would obtain that 
	$$j\in I_{\text{low}}(\bm{\alpha}^{k})\cap I_{\text{up}}(\bm{\alpha}^{k})\; \text{and}\; j\not\in I_{\text{low}}(\bm{\alpha}^{k+1}).$$
	
\noindent Since $i \in I_{\text{up}}(\bm{\alpha}^{k})$, we would have
	\begin{equation}
		\label{eq:s_k2}
		S_{1}(k+1)\leq S_{1}(k)\;\; \text{and}\;\; S_{2}(k+1)\leq S_{2}(k)-(N-j)+(N-i)\leq S_{2}(k)-1. 
	\end{equation}
The same reasoning (based on \eqref{eq:geij} and \eqref{eq:dim_6}) can be applied for any two consecutive iterations, from iteration $k$ to iteration $k+r$. By setting $r= N(N-1)$, \eqref{eq:s_k} and \eqref{eq:s_k2} imply that  $S_{1}(k)+S_{2}(k)$ is reduced to zero, which contradicts \eqref{eq:nocontr}. Therefore, \eqref{eq:violp} and \eqref{eq:delta} cannot hold are wrong and any limit point of the sequence satisfies the optimality conditions \eqref{eq:opt_cond}.

Since problem \eqref{eq:2BM}
	is strictly convex ($K \succeq 0$ and $C\sum\limits_{i \in I}G(\frac{\alpha_i}{C})- \lambda \sum\limits_{i \in I} \alpha_i$ is strictly convex), and its feasible set is compact, \eqref{eq:2BM} admits a unique global optimal solution $\bm{\alpha}^*$.
 
	From the previous arguments, all limit points of $\{\bm{\alpha}^k\}$ satisfy the optimality conditions, thus $\{\bm{\alpha}^k\}$ globally converges to $\bm{\alpha}^*$.
	
	\qed

\section{On hyperparameters tuning}
\label{subsec:c-lambda}
The values of hyperparameters $C$ and $\lambda$ can significantly impact the performance of the S-KLR method. Conducting a grid search to determine good estimates of these hyperparameters can be time-consuming and computationally expensive. \textcolor{black}{This appendix investigates the trend of testing accuracy and sparsity as functions of the two hyperparameters, considering the same training–testing split of the datasets reported in Section \ref{subsec:results}. Since we are interested in the overall impact of hyperparameter changes rather than in model comparison, the validation set used during tuning is not considered.} 

\begin{figure}[H]
\centering
\includegraphics{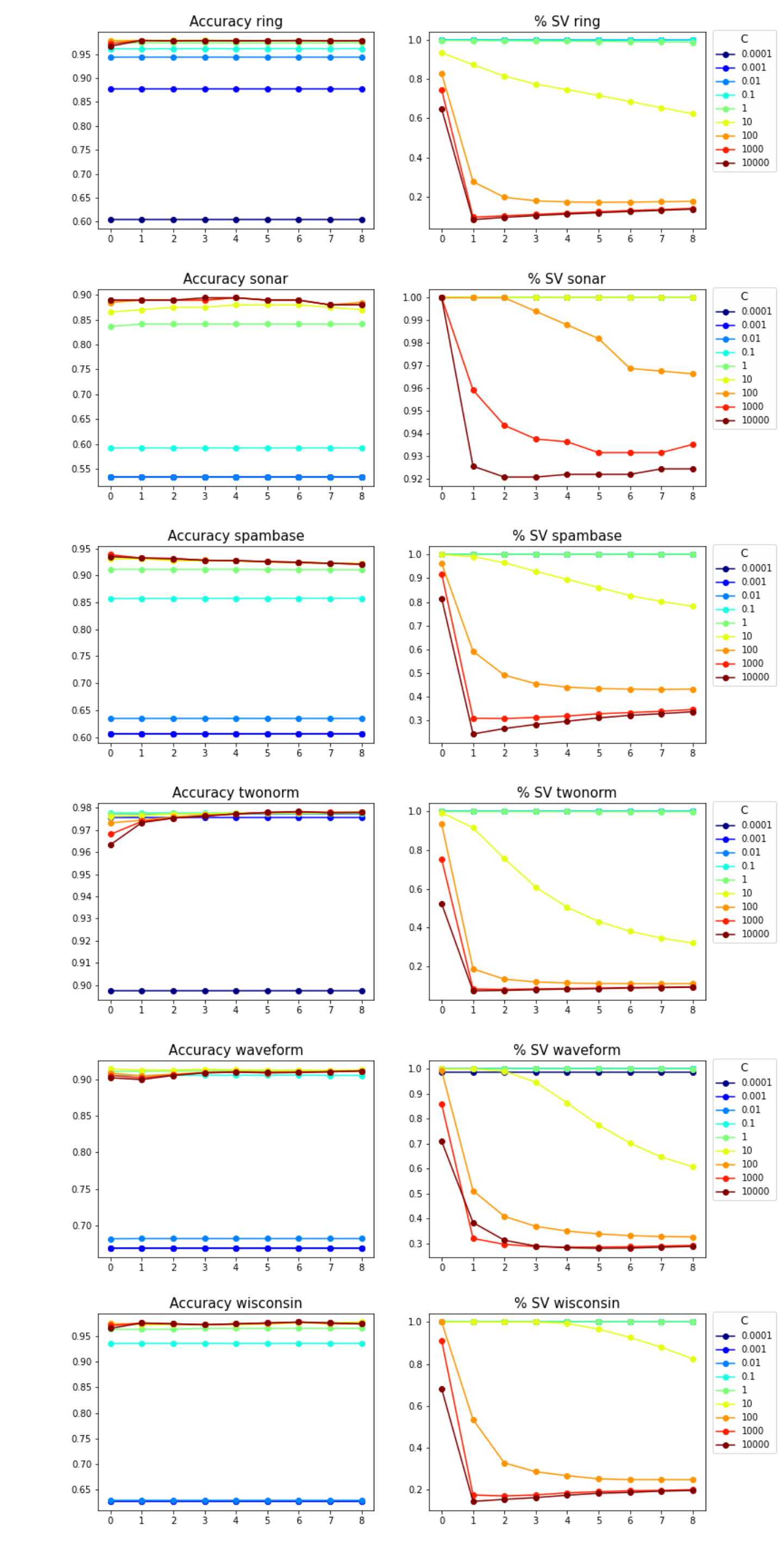}
    \caption{Testing accuracy and sparsity in terms of the regularization term $C$ and in the x-axis the sparsity term $\lambda$ for S-KLR (colored dotted line). The sparsity term $\lambda$ varies from $0$ to $C$ in 9 steps (i.e., 10 different values of $\lambda$).}
    \label{fig:plot_trend}
\end{figure}

Figure \ref{fig:plot_trend} highlights the impact on the performance of these hyperparameters, by showing the accuracy and sparsity trends obtained when varying $C$ and $\lambda$ according to the grid search described in Section \ref{subsec:results} for all considered datasets. In particular, the y-axis represents, respectively, the accuracy on the left-hand-side plots, and the sparsity level on the right-hand-side plots. The colors of the profiles represent the $C$ values (see the colorbar for values details), and the x-axis represent 10 equally values in the interval $\left[0,C\right]$ for $\lambda$.\par


We observed two main types of behaviors:
\begin{enumerate}
    \item[(i)] For a fixed value of $C$, the variation of $\lambda$ only slightly affect the accuracy (see plots on the left).
    \item[(ii)] For small values of $C$\footnote{In some sparsity plots the profiles related to small $C$ values (e.g., $C=0.0001$ or $C=0.001$) are overlapped by the ones associated to medium $C$ values (e.g., $C=1$ or $C=10$).}, the value of $\lambda$ does not significantly affect the model sparsity. However, for larger values of $C$, a small value of $\lambda$ is sufficient to sensibly increase the model sparsity. This behavior is consistent, with minor variations, across all 12 datasets.
\end{enumerate}
The observations (i) and (ii) suggest that the $\lambda$ value can be heuristically set to a relatively small value, without the need of a refined and expensive grid search.\par 
In the sequel, we set $\lambda=\frac{C}{10}$ and we refer to it as $\lambda$-choice.
 table reports the average results
(over the k = 5 folds) of the sparsest model among the $3$ most accurate ones.

In Table \ref{tab:result_eur}, we show the average results obtained with $\lambda$-choice and with S-KLR employing the standard grid search on both $\lambda$ and $C$. As in Table \ref{tab:first_results}, for both strategies we report the results corresponding to the best accuracy as well as the ones corresponding to the sparsest model among the $3$ most accurate ones. Concerning the best accuracy criterion, $\lambda$-choice yields comparable accuracy scores and a slight loss of average sparsity ($46.51\%$ against $41.64\%$). As to the sparsest among the $3$ most accurate criterion, $\lambda$-choice and S-KLR achieve comparable average performance for both accuracy and sparsity.  

To summarize, we observe that the simple $\lambda$-choice yields S-KLR models whose sparsity is not significantly worse than the one provided by S-KLR with a refined grid search for both $C$ and $\lambda$ values. Whereas the accuracy of both approaches is comparable. This shows that also in the case of KLR it is just sufficient to search for the value of the single hyperparameter ($C$) as it is the case of SVM, with no need for time consuming two dimensional grid search.
Considering, as an example, the grid search applied to the experiments of Section \ref{subsec:results}, $\lambda$-choice would have implied a computational time reduction by approximately a factor of 8 (the number of considered grid $\lambda$ values minus one).

\begin{table}[H]
\caption{Results in terms of testing accuracy and sparsity obtained by $\lambda$-heuristic ($(C,\lambda)$ set to $(C,\frac{C}{10})$) and by S-KLR. The standard deviation of the average accuracy is reported in parentheses. }\label{tab:result_eur}
\centering
\resizebox{\textwidth}{!}{
\begin{tabular}{lcccccccc}
\hline
& \multicolumn{2}{c}{$\lambda$-choice}       & \multicolumn{2}{c}{$\lambda$-choice}       & \multicolumn{2}{c}{S-KLR} & \multicolumn{2}{c}{S-KLR} \\ 
                                & \multicolumn{2}{c}{sparsest of 3 most accurate} & \multicolumn{2}{c}{}       & \multicolumn{2}{c}{sparsest of 3 most accurate} & \multicolumn{2}{c}{} \\ \cline{2-9} 
Dataset                         & acc   & $\%$ data points   & acc   & $\%$ data points            & acc    & $\%$ data points               & acc     & $\%$ data points  \\ \hline
\multicolumn{1}{l|}{banknote}   & 1     & \multicolumn{1}{c|}{0.059}& 1     & \multicolumn{1}{c|}{0.059} & 1      & \multicolumn{1}{c|}{0.05}  & 1       & 0.05          \\
\multicolumn{1}{l|}{coil2000}   & 0.94  & \multicolumn{1}{c|}{0.059} & 0.94  & \multicolumn{1}{c|}{0.059} & 0.94   & \multicolumn{1}{c|}{0.059} & 0.94    & 0.059         \\
\multicolumn{1}{l|}{diabetes}   & 0.768 & \multicolumn{1}{c|}{0.523} & 0.7695 & \multicolumn{1}{c|}{0.565} & 0.771  & \multicolumn{1}{c|}{0.567} & 0.771   & 0.567         \\
\multicolumn{1}{l|}{ionosphere} & 0.943 & \multicolumn{1}{c|}{0.488} & 0.949 & \multicolumn{1}{c|}{0.976} & 0.952  & \multicolumn{1}{c|}{0.496} & 0.952   & 0.496         \\
\multicolumn{1}{l|}{magic}      & 0.861 & \multicolumn{1}{c|}{0.357} & 0.861 & \multicolumn{1}{c|}{0.357} & 0.861  & \multicolumn{1}{c|}{0.357} & 0.865   & 0.99          \\
\multicolumn{1}{l|}{monk2}      & 0.958 & \multicolumn{1}{c|}{0.292} & 0.965 & \multicolumn{1}{c|}{1} & 0.97   & \multicolumn{1}{c|}{0.319} & 0.972   & 0.357         \\
\multicolumn{1}{l|}{ring}       & 0.978 & \multicolumn{1}{c|}{0.081} & 0.978 & \multicolumn{1}{c|}{0.081} & 0.978  & \multicolumn{1}{c|}{0.08}  & 0.978   & 0.08          \\
\multicolumn{1}{l|}{sonar}      & 0.889 & \multicolumn{1}{c|}{0.925} & 0.889 & \multicolumn{1}{c|}{0.925} & 0.894  & \multicolumn{1}{c|}{0.92}  & 0.894   & 0.92          \\
\multicolumn{1}{l|}{spambase}   & 0.932 & \multicolumn{1}{c|}{0.241} & 0.932 & \multicolumn{1}{c|}{0.241} & 0.932  & \multicolumn{1}{c|}{0.241} & 0.939   & 0.918         \\
\multicolumn{1}{l|}{twonorm}    & 0.974 & \multicolumn{1}{c|}{0.071} & 0.974 & \multicolumn{1}{c|}{0.185} & 0.977  & \multicolumn{1}{c|}{0.08}  & 0.978   & 0.082         \\
\multicolumn{1}{l|}{waveform}  & 0.902 & \multicolumn{1}{c|}{0.32} & 0.912 & \multicolumn{1}{c|}{0.99} & 0.911  & \multicolumn{1}{c|}{0.291} & 0.911   & 0.291         \\ 
\multicolumn{1}{l|}{wisconsin}  & 0.975 & \multicolumn{1}{c|}{0.144} & 0.975 & \multicolumn{1}{c|}{0.144} & 0.977  & \multicolumn{1}{c|}{0.187} & 0.977   & 0.187         \\ \hline
\multicolumn{1}{l|}{AVERAGE}    & 0.927 (0.064)  & \multicolumn{1}{c|}{0.297} & 0.929 (0.064)  & \multicolumn{1}{c|}{0.4651} & 0.9303 (0.064)  & \multicolumn{1}{c|}{0.304} & 0.9315 (0.064)   & 0.4164         \\ \hline
\end{tabular}
}
\end{table}

\begin{figure}[H]
    \centering
    \includegraphics{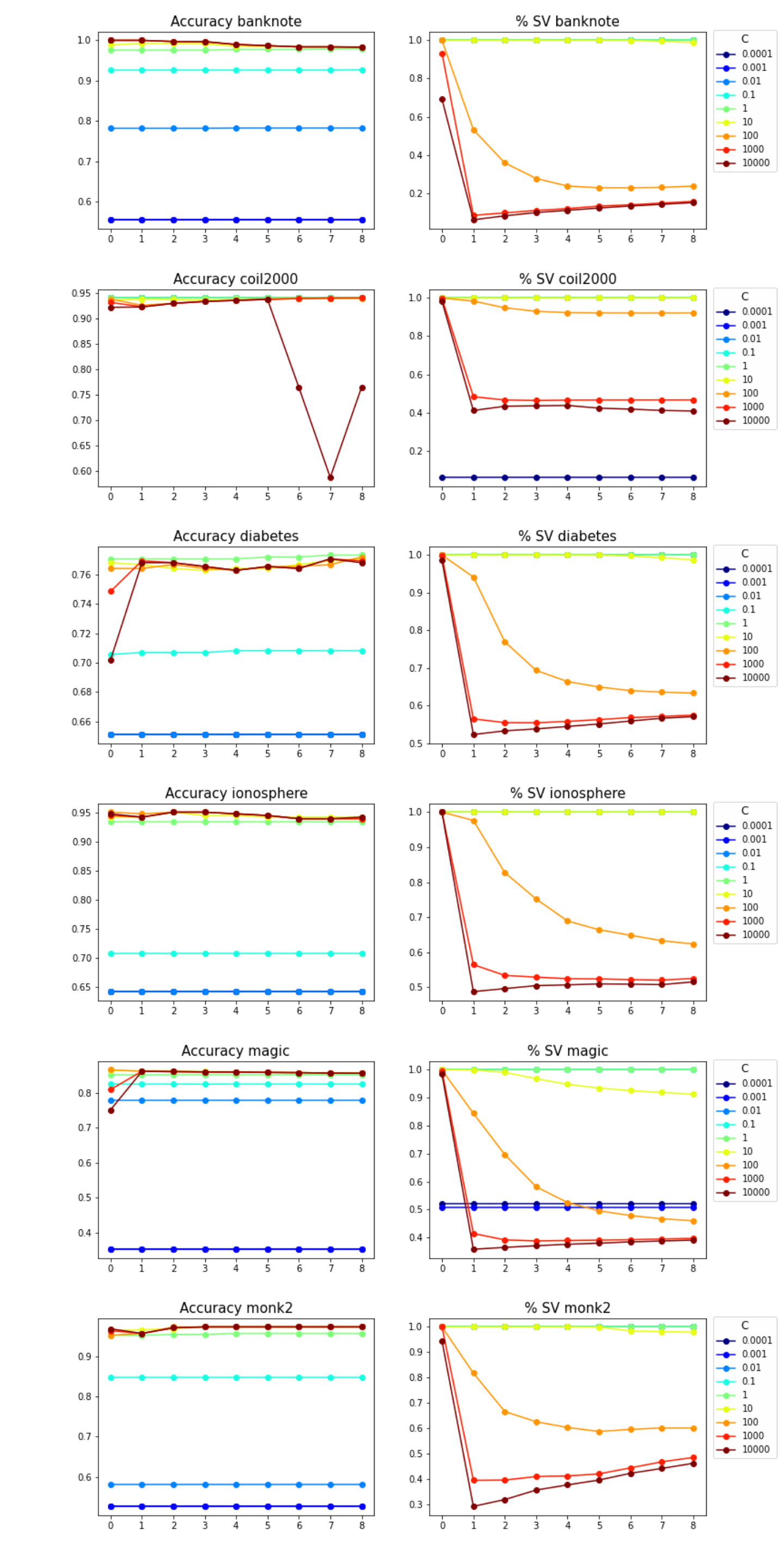}
    \end{figure}

\end{document}